\documentclass[lettersize,journal]{IEEEtran}
\usepackage{amsmath,amsfonts}
\usepackage{algorithmic}
\usepackage{algorithm}
\usepackage{array}
\usepackage[caption=false,font=normalsize,labelfont=sf,textfont=sf]{subfig}
\usepackage{textcomp}
\usepackage{stfloats}
\usepackage{url}
\usepackage{verbatim}
\usepackage{graphicx}
\usepackage{cite}
\usepackage{diagbox} 
\usepackage{cases}
\usepackage{multirow}
\usepackage{multicol}
\usepackage{adjustbox}
\usepackage{makecell}
\usepackage{xcolor}

\begin{document}

\title{CSI-BERT2: A BERT-inspired Framework for Efficient CSI Prediction and Classification in Wireless Communication and Sensing}

\author{Zijian Zhao,\IEEEmembership{} Fanyi Meng,\IEEEmembership{} Zhonghao Lyu,\IEEEmembership{} Hang Li\IEEEmembership{},\IEEEmembership{} Xiaoyang Li, Guangxu Zhu\IEEEmembership{}
\thanks{The work of Guangxu Zhu was supported in part by National Natural Science Foundation of China (Grant No. 62371313), in part by Guangdong Young Talent Research Project (Grant No. 2023TQ07A708), in part by Shenzhen-Hong Kong-Macau Technology Research Programme (Type C) (Grant No. SGDX20230821091559018), in part by the Shenzhen Science and Technology Program (Grant No. JCYJ20241202124934046). The work of Xiaoyang Li was supported in part by Young Elite Scientists Sponsorship Program by CAST under Grant YESS20240364, Shenzhen Science and Technology Program under Grants JCYJ20241202124934046 and KJZD20240903095402004. (Corresponding Author: Guangxu Zhu)}
\thanks{Zijian Zhao is with Shenzhen Research Institute of Big Data, Shenzhen 518115, China, and also with the School of Computer Science and Engineering, Sun Yat-sen University, Guangzhou 510275, China (e-mail: zhaozj28@mail2.sysu.edu.cn)}
\thanks{Fanyi Meng, Hang Li, and Guangxu Zhu are with the Shenzhen Research Institute of Big Data, The Chinese University of Hong Kong (Shenzhen), Shenzhen 518115, China (e-mail: fanyimeng@link.cuhk.edu.cn; hangdavidli@163.com; gxzhu@sribd.cn)}
\thanks{Xiaoyang Li is with the Department of Electrical and Electronic Engineering, Southern University of Science and Technology, Shenzhen, 518055, China (e-mail: lixy@sustech.edu.cn)}
\thanks{Zhonghao Lyu is with the Department of Information
Science and Engineering, KTH Royal Institute of Technology, Stockholm, Sweden (e-mail: lzhon@kth.se).}

}



\markboth{IEEE TRANSACTIONS ON MOBILE COMPUTING}%
{Shell \MakeLowercase{\textit{et al.}}: A Sample Article Using IEEEtran.cls for IEEE Journals}


\maketitle

\begin{abstract}

Channel state information (CSI) is a fundamental component in both wireless communication and sensing systems, enabling critical functions such as radio resource optimization and environmental perception. In wireless sensing, data scarcity and packet loss hinder efficient model training, while in wireless communication, high-dimensional CSI matrices and short coherent times caused by high mobility present challenges in CSI estimation.
\textcolor{black}{To address these issues, we propose a unified framework named CSI-BERT2 for CSI prediction and classification tasks, built on our previous work CSI-BERT, which adapts BERT to capture the complex relationships among CSI sequences through a bidirectional self-attention mechanism. We introduce a two-stage training method that first uses a mask language model (MLM) to enable the model to learn general feature extraction from scarce datasets in an unsupervised manner, followed by fine-tuning for specific downstream tasks. Specifically, we extend MLM into a mask prediction model (MPM), which efficiently addresses the CSI prediction task. To further enhance the representation capacity of CSI data, we modify the structure of the original CSI-BERT. We introduce an adaptive re-weighting layer (ARL) to enhance subcarrier representation and a multi-layer perceptron (MLP)-based temporal embedding module to mitigate temporal information loss problem inherent in the original Transformer.} 
Extensive experiments on both real-world collected and simulated datasets demonstrate that CSI-BERT2 achieves state-of-the-art performance across all tasks. Our results further show that CSI-BERT2 generalizes effectively across varying sampling rates and robustly handles discontinuous CSI sequences caused by packet loss—challenges that conventional methods fail to address.
\textcolor{black}{The dataset and code are publicly available at \url{https://github.com/RS2002/CSI-BERT2}.}


\end{abstract}


\section{Introduction}



Channel state information (CSI) plays a critical role in both wireless sensing and wireless communication systems by capturing the propagation characteristics of the wireless channel. This information provides valuable insights into the radio environment, including the channel conditions, signal quality, and the motion and location of objects. In wireless communication systems, CSI is essential for optimizing key processes in wireless communication systems, such as channel compensation, adaptive modulation and coding, user selection and scheduling. Effective utilization of CSI enables improved system performance, increased spectrum efficiency, and more reliable communication, ultimately enhancing the overall user experience \cite{zhu2023pushing,li2023integrated,wen2023task}. In wireless sensing applications, CSI can enhance the ability to detect and track objects, leading to improved environmental awareness \cite{xu2023edge,yao2024wireless,yang2025privacy}. Owing to its advantages in privacy protection, low cost, and penetration ability, wireless sensing has been widely adopted in diverse applications, such as fall detection \cite{cai2023falldewideo}, localization \cite{lofi}, gesture recognition \cite{zhao2025does}, and person identification \cite{CSi-Net}.


Despite the importance of CSI in wireless systems, acquiring, processing, and utilizing CSI data poses significant challenges. In wireless communication, estimating CSI is time-consuming, particularly in high-dimensional and high-mobility scenarios. Frequent updates in rapidly changing environments increase computational complexity. Consequently, delays in obtaining accurate CSI can degrade spectrum efficiency by leading to suboptimal resource allocation and increased interference.
In wireless sensing, packet loss due to factors such as channel noise and device errors may severely impact model performance. As noted in \cite{CSI-BERT}, even in relatively simple indoor environments, the average packet loss rate can reach 14.5\%, with peaks up to 70\% within a single second. Such discontinuities in CSI sequences pose serious challenges for the design and deployment of wireless sensing models.
Additionally, for data-driven methods in both scenarios, a major challenge is the limited availability of datasets. Collecting large datasets independently is costly, while public datasets are often hetergeneous and scarce, making them difficult to use in combination. Consequently, how to enable models to learn effectively from limited datasets has become an important research topic.

Recent studies have noted and addressed these challenges. In wireless communication, the high computational complexity of CSI estimation has driven the adoption of deep learning techniques to predict future CSI more efficiently. However, most existing approaches rely on auto-regressive (AR) manners \cite{LSTM_CSI1,LSTM_CSI2,LSTM_CSI3,LSTM_CSI4}, which have been criticized for error accumulation \cite{parthipan2024defining}.
Regarding the packet loss problem in wireless sensing, most previous methods employ interpolation techniques such as linear interpolation and Kriging to estimate CSI within each dimension, often neglecting the inter-dependencies among subcarriers. Prior research \cite{CSI-BERT} has indicated that these interpolation methods provide only limited benefits for sensing models.
To address the issue of data scarcity, the pre-training and fine-tuning paradigm has been extensively explored in wireless systems and has shown promising performance \cite{lyu2024rethinking}. By pre-training on either downstream datasets or unlabeled datasets, the model can achieve a more generalized capacity for data representation, thereby enhancing its generalization and robustness in downstream tasks.\footnote{To avoid confusion, we clarify the concept of ``pre-training" in this paper. We use this term in its original and broadest sense: any initial training phase that improves model performance on downstream tasks, regardless of the data source (labeled/unlabeled, in-domain/out-of-domain) or methodology (self-supervised, supervised, etc.). This intentionally differs from narrower interpretations that restrict ``pre-training" exclusively to large-scale unsupervised learning on massive datasets, as commonly employed in today's large models.}
Finally, given the wide diversity of tasks in wireless communication and sensing, designing specialized models for each task can be labor-intensive and inefficient. A unified and generic framework is therefore highly desirable \cite{foundation_survey}, and many works \cite{yang2025wirelessgpt,liu2025wifo,jiang2025mimo} have begun exploring the development of foundational models for wireless systems.

To address the aforementioned shortcomings, we propose a method inspired by bidirectional encoder representations from transformers (BERT) \cite{BERT}, named CSI-BERT2, which is an improved version of our previous work, CSI-BERT \cite{CSI-BERT}. In CSI-BERT, the masked language model (MLM) approach was first introduced for the pre-training of the CSI recovery model in Wi-Fi sensing, enabling it to learn effectively from limited dataset and improving performance in downstream tasks.
Building on this, CSI-BERT2 introduces the mask prediction model (MPM), where we mask the final portion of the CSI sequence and train the model to recover it. By predicting the masked CSI in a single step, our method efficiently avoids the error accumulation problem common in AR models.
Additionally, we propose a novel multi-layer perceptron (MLP)-based time embedding method that effectively addresses the \textcolor{black}{temporal information loss problem \cite{transformer-time} inherent in the original Transformer. As illustrated in \cite{transformer-time}, the permutation-invariant nature of the attention mechanism leads to temporal information loss, even with positional embeddings. In contrast, our proposed time embedding method explicitly exploits the temporal order, which is critical for CSI sequences, thereby mitigating this issue.}
This design enables CSI-BERT2 to generalize efficiently in environments with varying sampling rates, and perform robustly even in a zero-shot setting.
Finally, considering the varying information content convened across different subcarriers, we introduce an adaptive re-weighting layer (ARL) \cite{arl}, which helps to adaptively capture the importance of each subcarrier. Through these architectural enhancements, we successfully address the low performance of the original CSI-BERT in CSI classification tasks.
In conclusion, the main contributions of this paper are summarized as follows:

\noindent \textbf{(1) A Multifunctional Framework for CSI Time Series:} We propose CSI-BERT2, a multifunctional model for CSI time series tasks in wireless systems, including CSI prediction and CSI classification. The model supports easy adaptation to various tasks simply by replacing the classification or prediction head. To realize these, we optimize the structure and training process of the original CSI-BERT model, expanding its application from solely CSI recovery to include both CSI classification and prediction.

\noindent \textbf{(2) Improved Model Structure:} We design a novel time embedding technique that utilizes MLP to further encode temporal information, addressing the temporal information loss problem inherent in the Transformer positional embedding layer. This design enables our model to function effectively in various sampling rate scenarios. Furthermore, we propose a simple attention mechanism called ARL to better capture the relationships and importance between subcarriers adaptively.

\noindent \textbf{(3) Improved Training Process:} To make full use of the limited training data, we propose a two-stage training strategy. Specifically, we first pre-train the model to learn general data representations using unsupervised MLM, followed by fine-tuning for specific downstream tasks. Additionally, we expand MLM to the MPM, where only the last portion of CSI sequences is masked, enabling the model to effectively perform CSI prediction tasks.

\noindent \textbf{(4) Experiment Evaluation:} To validate the performance of CSI-BERT2, we conduct comprehensive experiments across four datasets, including both public and self-collected data from real-world and simulated environments. 
In the CSI prediction task, CSI-BERT2 achieves the lowest error, outperforming the current state-of-the-art (SOTA) by over 60\% in mean squared error (MSE). Although it requires slightly more computation time than other models, the total prediction time for the entire dataset, consisting of hundreds of samples, remains under 0.5 seconds, demonstrating its practicality for real-time deployment.
In the CSI classification task, CSI-BERT2 consistently achieves the highest accuracy across various applications, including gesture recognition, person identification, fall detection, and crowd counting. Notably, our method shows strong robustness to discontinuous data caused by packet loss and to sampling rate mismatches between the training and testing sets, without the need to process and recover data or train a new model separately. Its accuracy drops by no more than 2\%, in contrast to other models that suffer tens of percentage points in degradation. By demonstrating robust performance under packet loss and variable sampling rates, CSI-BERT2 eliminates the need for resource-intensive retraining in dynamic environments, making it viable for real-time applications such as emergency fall detection.



The structure of this paper is organized as follows. Section \ref{Problem Statement} introduces the fundamental principles of CSI, the background on CSI prediction and classification, and a review of related works. Section \ref{Methodology} provides a detailed description of the model structure and the training process for different tasks. Section \ref{Experiment} presents a series of comparative experiments to demonstrate the advantages of our method, and several ablation studies to illustrate the efficiency of our design. Finally, Section \ref{Conclusion} summarizes the paper and highlights potential directions for future research.



\section{Preliminaries and Related Works}
\label{Problem Statement}
\subsection{CSI Basics}

Following \cite{ma2019wifi}, in Wi-Fi communication,
CSI characterizes the way Wi-Fi signals propagate from the transmitter (TX) to the receiver (RX). Modern Wi-Fi systems typically employ orthogonal frequency division multiplexing (OFDM) as the modulation technique. OFDM divides the wideband channel into multiple narrowband subcarriers in the frequency domain, each experiencing flat fading for easier processing. In the time domain, data is transmitted as a series of OFDM symbols with a cyclic prefix to prevent interference from multipath delays. The amplitude and phase of CSI are influenced by multipath effects, which include amplitude attenuation and phase shift. Each element of the CSI matrix $H$ is given by:
\begin{equation}
\begin{aligned}
H(f; t) = \sum_{n=1}^{N} a_n(t) e^{-j2\pi f \tau_n(t)} \ ,
\label{csi}
\end{aligned}
\end{equation}
where $N$ is the number of paths between the TX and the RX, \( a_n(t) \) is the amplitude attenuation factor, \( \tau_n(t) \) is the propagation delay, \( f \) is the carrier frequency, and \( t \) is the timestamp.

\subsection{CSI Prediction}

Obtaining accurate CSI matrix in wireless communication remains a challenging task \cite{qi2021acquisition}. The large number of antennas results in a high-dimensional CSI matrix, which in turn makes the channel estimation algorithms difficult to solve. Additionally, when objects in the environment move at high speeds, the channel coherence time shortens considerably, imposing stricter requirements on the channel estimation speed. As illustrated in Fig. \ref{CSI prediction}, CSI plays a critical role in optimizing wireless communication system  performance, making fast and reliable CSI estimation essential. Furthermore, predicting future CSI can provide valuable opportunities for system-level optimizations, such as effective channel compensation.

Recently, some works have explored the use of additional information for CSI prediction. For example, in \cite{9277535}, the authors used multimodal sensory data, including received pilots, user position, and historical downlink CSI to predict the current CSI. In \cite{zhang2022predicting}, a deep learning-based framework using a three dimensional (3D) CNN architecture effectively predicts downlink CSI by learning temporal, spatial, and frequency correlations from past channel sequences. The authors in \cite{8815557} have proposed an machine learning (ML)-based architecture for estimating channels by using CNN-AR, which achieves gains in prediction quality.

\subsection{CSI Classification}

In wireless sensing, many tasks can be formulated as CSI classification problems, including gesture recognition\cite{widar}, fall detection \cite{cai2023falldewideo}, and activity recognition \cite{zhang2023ratiofi}. Reviewing CSI basics, human movement in the environment can cause changes in the multipath channel, which in turn changes the extracted CSI matrix. Therefore, by analyzing the pattern of changes in CSI, it is possible to sense human activities. 

However, in real-world sensing systems, accurately characterizing human actions are challenging. Complex actions, such as falling, kicking, and squatting, are particularly difficult to model mathematically. Additionally, noise in Wi-Fi extraction devices, including carrier frequency offset and sampling time offset, is hard to completely eliminate. These challenges motivate the use of deep learning techniques for Wi-Fi sensing. For example, in \cite{metafi}, the authors have estimated human poses from videos to supervise the training of Wi-Fi CSI. The broad learning system proposed in \cite{fingerprint} aims to achieve fast and accurate Wi-Fi fingerprint localization. Furthermore, \cite{wang20223d} have proposed Wi-Mesh, which utilizes Wi-Fi signals and 2D angle-of-arrival (AoA) estimation to create 3D human meshes using deep learning, showing robustness across various environments and conditions.

Nonetheless, unstable network conditions often lead to signal demodulation failures at the receiver due to factors, such as weak signal strength, frequency interference, and hardware errors. Moreover, packet loss itself may contain valuable sensing information. For instance, when an object obstructs the line-of-sight (LoS) path, the signal strength at the receiver may significantly decreases, leading to an increase in packet loss. These discontinuities in CSI resulting from packet loss can severely degrade the performance of wireless sensing models. Unfortunately, this issue has not been well addressed in existing literature. Most studies attempt to use linear interpolation when the packet loss rate is low \cite{9645160}. However, traditional interpolation methods fail under high loss rates, resulting in biased parameter estimation and incorrect inference.

To tackle this challenge, our previous work have introduced CSI-BERT \cite{CSI-BERT}, a self-supervised learning method designed to capture the internal structure of CSI data and recover lost packets. The recovered CSI was shown to substantially enhance the performance of downstream sensing tasks. In this work, we further optimize the structure of CSI-BERT to improve its ability to capture temporal information, enabling it to successfully perform CSI classification tasks using the discontinuous CSI for training.

\section{Methodology} \label{Methodology}

\subsection{Overview}

In this section, we will introduce the data preprocessing methods, neural network architecture, pre-training process, and the fine-tuning and inference methods of CSI-BERT2 in turn. As a continuation of CSI-BERT, we first provide a brief introduction to it. (This section will primarily focus on the modified and enhanced components, while more details about CSI-BERT can be found in our previous paper \cite{CSI-BERT}.)

Prior to CSI-BERT, several works had already applied BERT \cite{BERT} to wireless sensing tasks such as people localization \cite{BERT-loc1,BERT-loc2} and radio map construction \cite{BERT-map}, based on the intuition that the bidirectional encoder of the Transformer \cite{Transformer} can effectively capture the relationships among packets by the self-attention mechanism. However, these studies typically fed discrete signals into the original BERT embedding layer, which leads to significant information loss. In contrast, CSI-BERT first proposed a novel embedding layer tailored to the structure of CSI data, where a multi-layer perceptron (MLP) is used to encode continuous CSI data directly. Additionally, a positional-embedding style time-embedding layer was introduced to capture the temporal information among packets. Inspired by the similarity between packet recovery and masked language modeling (MLM), CSI-BERT successfully achieves high-quality packet recovery through unsupervised learning without labeled data. The recovered CSI data enhances the performance of other sensing models in classification tasks, outperforming traditional interpolation methods that fail to consider the internal relationships among subcarriers.

However, the original CSI-BERT does not fully leverage the advantages of MLM, resulting in subpar performance in downstream tasks. For example, the classification accuracy of CSI-BERT after fine-tuning does not surpass that of the vanilla ResNet18 \cite{Resnet}. This limitation can be attributed to the inadequacies of the embedding layer. To address this issue, CSI-BERT2 introduces an ARL-based spatial encoding module to capture the relationships and importance among subcarriers. Additionally, a MLP-based temporal encoding module is proposed to capture both the order and distance relationships among packets, effectively solving the temporal information loss problem inherent in conventional positional embedding modules, which capture distance relationships without considering order information. Moreover, to fully utilize the learned knowledge during pre-training, we propose a novel fine-tuning method named mask prediction model (MPM) for the CSI prediction task, wherein the last portion of the CSI sequence is randomly masked, and the model is trained to predict these masked values. Compared to conventional AR-based methods, our approach can complete the prediction in one step, thereby eliminating the error accumulation problem.

\begin{figure}[htbp]
\centering 
\includegraphics[width=0.45\textwidth]{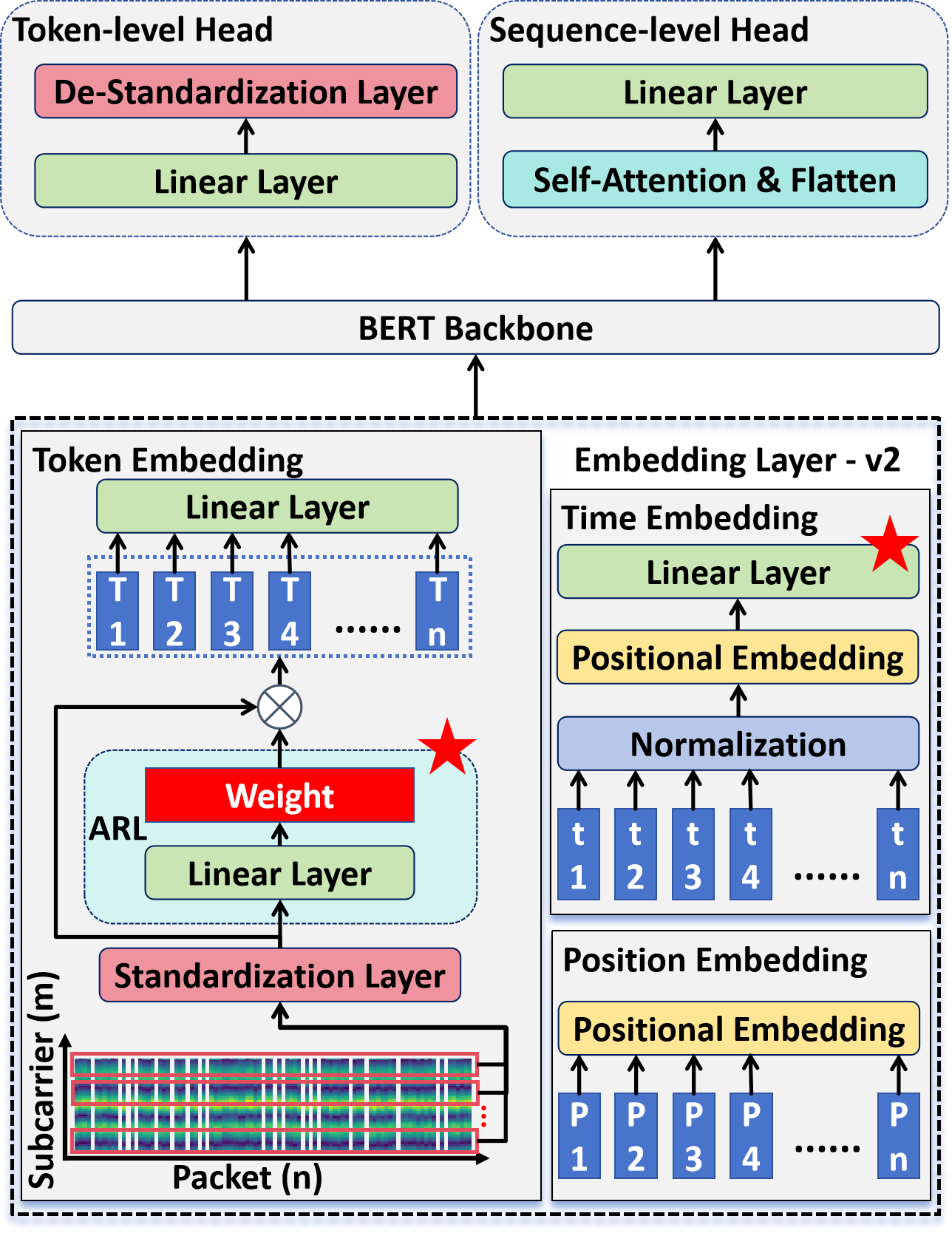}
\caption{CSI-BERT2 network architecture: In the figure, `T' represents token, `t' denotes timestamp, and `P' indicates the position. The red star highlights the different components of the CSI-BERT.}
\label{CSI-BERT2}
\end{figure}

\begin{figure}[h!]
\centering 
\subfloat[CSI Recovery \label{CSI recover}]{\includegraphics[width=0.39\textwidth]{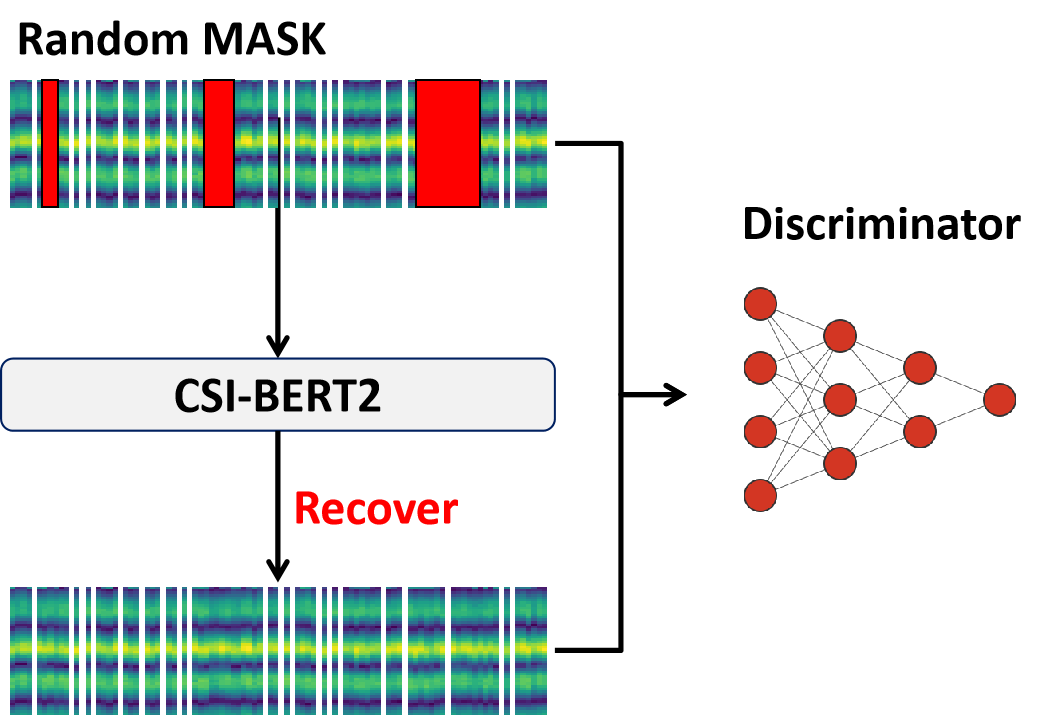}} \\
\subfloat[CSI Prediction \label{CSI prediction}]{\includegraphics[width=0.42\textwidth]{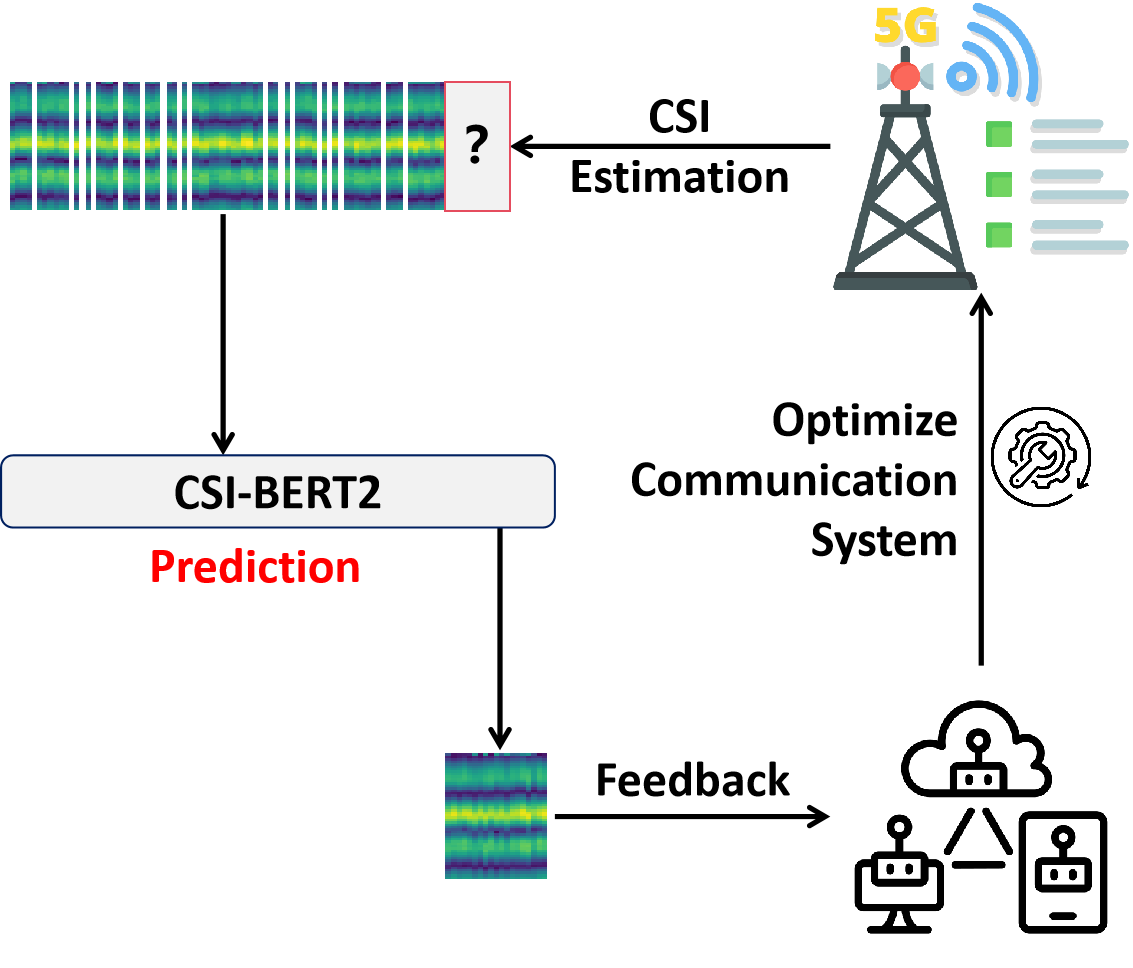}} \\
\subfloat[CSI Classification \label{CSI classification}]{\includegraphics[width=0.47\textwidth]{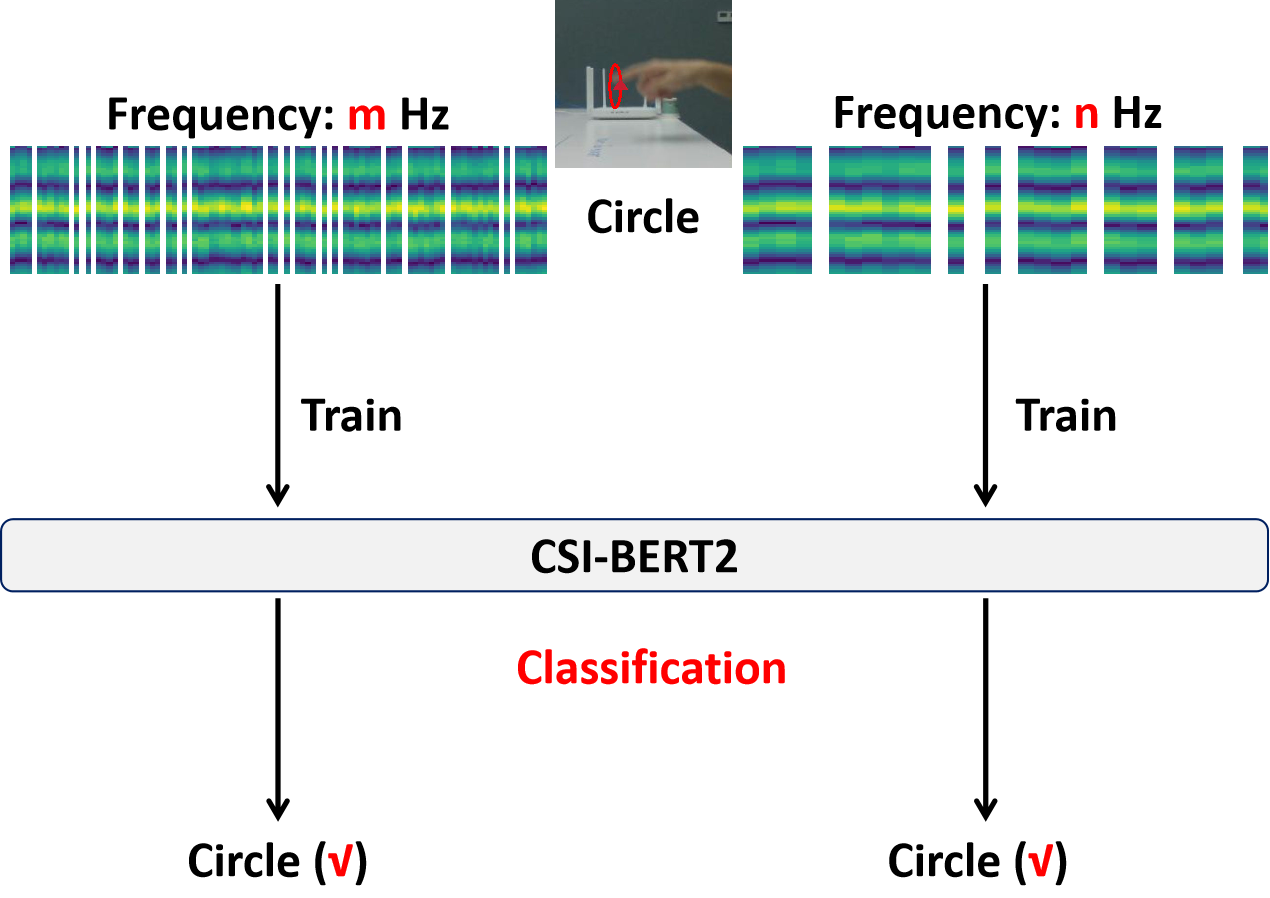}}
\caption{Three tasks in this paper: (a) During the unsupervised training phase, we randomly mask some CSI tokens and train the CSI-BERT2 to recover them in an unsupervised manner, while a discriminator is employed to enhance the realism of the recovered results. (b) The CSI prediction task focuses on swiftly predicting future CSI series for proactive communication optimization. (c) The CSI classification task utilizes CSI for specific sensing applications, such as gesture recognition. Notably, our CSI-BERT2 can efficiently process CSI data across varying sampling rates, a shortcoming for most other models.}
\label{task}
\end{figure}

The model structure and working scenarios of CSI-BERT2 are illustrated in Fig. \ref{CSI-BERT2} and Fig. \ref{task}, respectively. \textcolor{black}{We highlight the differences between CSI-BERT and CSI-BERT2 in Fig. \ref{CSI-BERT2} and Table \ref{tab:compare}.} We leverage the CSI sequence $c = [c_1, c_2, \ldots, c_N]$ and the corresponding timestamps $t = [t_1, t_2, \ldots, t_N]$ as input. Each $c_i$ represents a flattened vector of the CSI matrix at time $t_i$, which contains the amplitude and phase information of the subcarriers.


\begin{table}[htbp]
\caption{\textcolor{black}{Comparison between CSI-BERT \cite{CSI-BERT} and CSI-BERT2: Here PE represents the original positional embedding in Transformer, and SFT represents supervised fine-tuning. We utilize \underline{underline} to highlight the differences between CSI-BERT2 and CSI-BERT.}}
\begin{adjustbox}{width=0.49\textwidth}
    \centering
    \begin{tabular}{|c||c|c|}
        \hline
        & \textbf{CSI-BERT \cite{CSI-BERT}} & \textbf{CSI-BERT2} \\
        \hline \hline
        & \multicolumn{2}{c|}{\textbf{Model Structure}} \\
        \hline
        \textbf{Spatial Embedding} & MLP & MLP + \underline{ARL} \\
        \hline 
        \textbf{Temporal Embedding} & PE & PE + \underline{MLP} \\
        \hline 
        \textbf{Positional Embedding} & PE & PE \\
        \hline 
        \textbf{Backbone} & BERT & BERT \\
        \hline 
        \textbf{Discriminator} & DANN-based & \underline{GAN}-based \\
        \hline
        & \multicolumn{2}{c|}{\textbf{Training Method}} \\
        \hline
        \textbf{Pre-training} & MLM & MLM \\
        \hline
        \textbf{Classification Fine-tuning} & SFT & SFT + \underline{Random Mask} \\
        \hline
        \textbf{Prediction Fine-tuning} & $\times$ & \underline{MPM} \\
        \hline
        & \multicolumn{2}{c|}{\textbf{Application Scenarios}} \\
        \hline
        \textbf{CSI Recovery} & $\checkmark$ & $\checkmark$ \\
        \hline
        \textbf{CSI Classification} & Single Sampling Rate & \underline{Various} Sampling Rates \\
        \hline
        \textbf{CSI Prediction} & $\times$ & \underline{$\checkmark$} \\
        \hline
    \end{tabular}
\end{adjustbox}
\label{tab:compare}
\end{table}

\subsection{Data Pre-processing}

Due to packet loss, the received CSI series are always incomplete. Before inputting them to our model, we need to first identify where the packet loss happens and use a placeholder [PAD] to occupy the place of lost packets. 
We determine the occurrence of packet loss by analyzing the time gap between two successive timestamps. 
Given a sampling rate $f$, the ideal time gap between two successive timestamps should be $\Delta t = \frac{1}{f}$ seconds. Assume we receive a CSI series $[c_1, c_2, \dotsc, c_n]$ with corresponding timestamps $[t_1, t_2, \dotsc, t_n]$ with the unit of seconds. Then we can identify whether there is packet loss and how many packets were lost between $c_i$ and $c_{i+1}$. We calculate the lost packet amount according to:
\begin{equation}
k = \max \{\text{round}\left(\frac{t_{i+1} - t_i}{\Delta t} -1\right) ,0\} \ .
\label{package loss}
\end{equation}
Then we will add $k$ [PAD] tokens between $c_i$ and $c_{i+1}$. Besides, we also need to add $k$ corresponding timestamps between $t_i$ and $t_{i+1}$. We define the $j-\text{th}$ added timestamp between $t_i$ and $t_{i+1}$ as:
\begin{equation}
t_{i,j} = t_i + \frac{t_{i+1} - t_i}{k} j + \epsilon_j \ ,
\label{timestamp inter}
\end{equation}
where $\epsilon_j$ is a small random number used to simulate the real-time gap. Our method to identify packet loss is reasonable because after adding the [PAD] tokens, we found the packet amount is very close to the theoretical amount when there is no packet loss in our dataset.

\subsection{Model Structure}
As shown in Fig. \ref{CSI-BERT2}, we utilize BERT \cite{BERT} as the backbone of CSI-BERT2, while replacing the bottom embedding layer and the top heads to accommodate our data format and tasks. The embedding layer consists of three components, which can be divided into two parts: token embedding for spatial encoding, and time embedding along with position embedding for temporal encoding. For the top heads, our design is based on the commonly used dual-branch head in Transformers, which includes a sequence-level head (where the entire sequence produces one output) and a token-level head (where each token generates one output).

\subsubsection{Spatial and Temporal Embedding Layer}

To feed CSI data to the BERT backbone, we use an MLP-based method to extract features within each packet. However, in CSI data, different subcarriers contain varying amounts of information and importance. For example, guard carriers remain relatively consistent across packets, providing limited guidance for downstream tasks. Consequently, we aim for the model to learn the importance and relationships among subcarriers autonomously. The ARL module \cite{arl} is an effective choice, as it can adaptively assign different weights to different channels (subcarriers). The principle of ARL can be represented as:
\begin{equation}
\begin{aligned}
T = \text{MLP}(c) \cdot c \ ,
\label{ARL}
\end{aligned}
\end{equation}
where $c$ is the input, and $T$ is the output token. ARL is a mechanism similar to attention but simpler and more efficient. It first employs an MLP to generate a weight vector and then multiplies this vector with the original input. This mechanism ensures that the network assigns greater importance to subcarriers containing informative content while down-weighting those with limited variation, such as guard carriers. As a result, it enhances the model’s precision and overall effectiveness. Subsequently, an additional MLP is applied along the subcarrier dimension to further extract features.

Additionally, prior to inputting the data, we follow CSI-BERT by using a standardization layer over the time (packet) dimension. This design is motivated by the observation that the distribution of CSI in each subcarrier changes significantly over time. Therefore, the time-dimension standardization operation can help the model learn useful features more effectively and increase its generalization capacity by mitigating the covariate shift problem \cite{KNN-MMD,covariate_shift}. The operation can be represented as:
\begin{equation}
\begin{aligned}
& \mu^{(j)}_n=\frac{\sum_{n=1}^N c_n^{(j)}}{N} \ ,\\
& \sigma^{(j)}_n=\sqrt{\frac{\sum_{n=1}^N (c_n^{(j)}-\mu_n^{(j)})^2}{N}} \ ,\\
& \text{Standard}(c_n^{(j)})=\frac{c_n^{(j)}-\mu_n^{(j)}}{\sigma_n^{(j)}} \ ,
\label{standard}
\end{aligned}
\end{equation}
where $c_n^{(j)}$ represents the $j$-th dimension of $c_n$, and $\mu$ and $\sigma$ denote the mean and standard deviation, respectively.

Compared to the spatial encoding, which primarily focuses on the inner information within each packet, the temporal encoding aims to provide more order and position information to the BERT backbone for better learning of the relationships among packets. However, we noted that the vanilla positional embedding in Transformers is insufficient. First, the arrival gap between successive packets is not uniform due to various factors such as environmental noise and movement. The positional embedding cannot capture these small differences. Most importantly, it has been shown to be weak in handling time series data \cite{transformer-time}, which can lead to temporal information loss.

To address these issues, we propose using an MLP to capture time information more effectively:
\begin{equation}
\begin{aligned}
& \text{TE}(t_i)= \text{MLP}( \text{PE}(t_i) ) \ , \\
& \text{PE}(t_i)^{(j)} =
    \begin{cases}
        \text{sin}\left(\frac{\text{Norm}(t_i)}{10^\frac{4j}{d}}\right), \quad j=2k \\
        \text{cos}\left(\frac{\text{Norm}(t_i)}{10^\frac{4(j-1)}{d}}\right), \quad j=2k+1
    \end{cases}
\ , \\
& \text{Norm}(t_i)=\frac{t_i-\text{min}(t)}{\text{max}(t)-\text{min}(t)} L \ ,
\label{time embedding}
\end{aligned}
\end{equation}
where TE and PE represent the time embedding and positional embedding, respectively, $\text{PE}(t_i)^{(j)}$ denotes the $j$-th dimension of $\text{PE}(t_i)$, $k$ is a positive integer, $L$ is the maximum length of the CSI sequence, and $d$ is the dimension of each CSI vector $c_i$. We first normalize the timestamp following the positional embedding method to encode it, and then feed it into the MLP for more complex information encoding. In this way, the original positional embedding captures the relative position information, while the time embedding encodes absolute temporal positions.

\subsubsection{Dual-branch Head Layer}
CSI-BERT2 adopts a dual-branch architecture to support different tasks. The token-level head is used for CSI recovery and prediction tasks, while the sequence-level head is used for CSI classification tasks.

\begin{figure}[htbp]
\centering 
\subfloat[Average]{\includegraphics[width=0.25\textwidth]{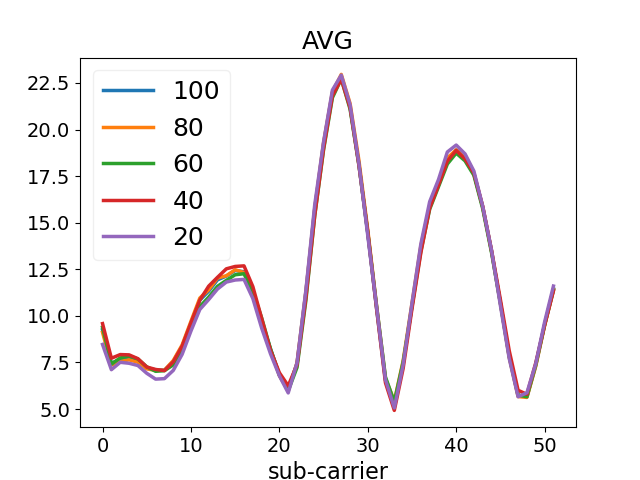}} 
\subfloat[Standard Deviation]{\includegraphics[width=0.25\textwidth]{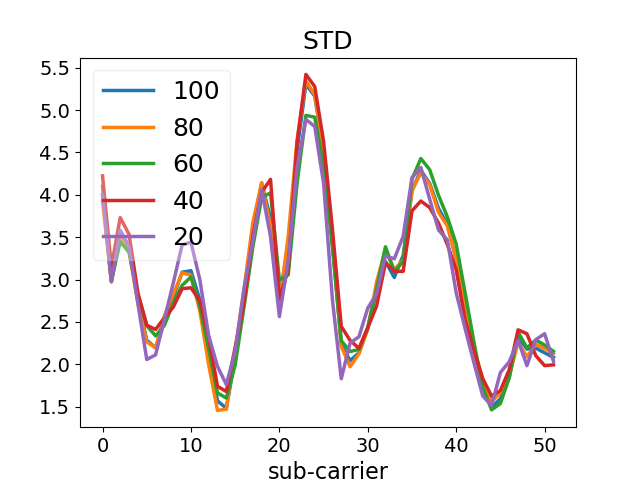}} \\
\caption{Average and standard deviation of CSI amplitude (WiGesture dataset \cite{CSI-BERT}): Each line represents a different number of packets sampled from one second. The x-axis represents the subcarriers.}
\label{fig:Wigesture Analysis}
\end{figure}

For the token-level head, a de-standardization layer is used to ensure that the output has a similar distribution as the input:
\begin{equation}
\begin{aligned}
    \text{De-Standard}(y_i^{(j)})=(y_i^{(j)}+\mu_i^{(j)})*\sigma_i^{(j)} \ , 
\label{de-standard}
\end{aligned}
\end{equation}
where $y$ represents the output of the last linear layer, $\mu$ and $\sigma$ are calculated using Eq. (\ref{standard}). The considered de-standardization layer is both necessary and efficient. This is because, on the one hand the distribution information is discarded by the standardization layer, we can only re-introduce this information by the de-standardization layer. On the other hand the mean and standard deviation of CSI are relatively stable within a small time window. For example, in Fig. \ref{fig:Wigesture Analysis}, we use the WiGesture dataset \cite{CSI-BERT} to illustrate the average and standard deviation of CSI amplitude for each subcarrier in one second. We randomly select a subset of packets from a total of 100, and show their average and standard deviation. The result shows that even with a change in the number of packets, the two statistics do not change significantly. It also implies that regardless of packet loss rate, these two statistics remain nearly consistent within a one-second interval. As a result, the de-standardization operation can help the model obtain a relatively realistic distribution information in both recovery task and prediction task.

\color{black}
For the sequence-level head, we use the same architecture as most Transformer-based methods \cite{yang2016hierarchical,MidiBERT}, which first incorporates a modified self-attention layer to combine the features in the time dimension and then uses a linear layer to generate the final output:
\begin{equation}
\begin{aligned}
x &  \in \mathbb{R}^{N,d_1} \ , \\
M_{attn} & = \text{Softmax}(\text{MLP}(x)) \in \mathbb{R}_+^{N,d_2} \ , \\
z & = \text{Flatten}(M_{attn}^T \cdot x) \in \mathbb{R}^{d_1 \times d_2} \ , \\
y & = \text{MLP}(z) \in \mathbb{R}^{o} \ ,
\end{aligned}
\end{equation}
where $d_1$ and $d_2$ represent the feature dimensions, $o$ denotes the output dimension, and $x$, $z$, and $y$ correspond to the input, latent feature, and output, respectively.
\color{black}

\textcolor{black}{Finally, considering the significance of real-time CSI classification and prediction models, we further analyze the time complexity of CSI-BERT2. The method primarily consists of self-attention and linear layers, so the complexity can be approximately expressed as $O(\alpha N^2 d + \beta N d^2)$, where $N$ is the sequence length, $d$ represents the feature dimension, $\alpha$ is the number of self-attention layers, and $\beta$ is the number of linear layers. For simplification, we neglect the dimensional differences between the input, output, and hidden layers. Other computations, such as positional embedding encoding, dot products in ARL, and activation functions, can be absorbed by the two main terms. The self-attention complexity mainly arises from the BERT backbone, while the linear layers come from the BERT backbone, the embedding layer, and the output heads.}

\subsection{Unsupervised Pre-training Process}

In the pre-training phase of CSI-BERT2, we employ MLM to enable the model to learn basic data features in an unsupervised manner. Additionally, an adversarial learning strategy is also implemented to assist the model in recovering more realistic CSI data. Compared to CSI-BERT, we replace the adversarial learning approach of domain adversarial neural network (DANN) \cite{DANN} with generative adversarial networks (GAN) \cite{GAN}. This change is motivated by the observation that the performance of DANN heavily relies on the careful tuning of the gradient reversal coefficient $\lambda$, which can vary dynamically across different training conditions. Additionally, we found that a fixed $\lambda$ may not perform consistently when the training data changes, potentially causing the model to diverge. In contrast, the GAN-based approach offers greater stability. In the following sections, we provide the detailed introduction to the pre-training process.

For CSI sequence, the MLM involves filling empty positions with [PAD] tokens, randomly masking tokens with [MASK] tokens, and training the model to recover the masked tokens. In BERT, the value of [MASK] token is fixed, but in CSI-BERT2, we replace the word embedding layer with a linear layer, which means the [MASK] token can take on different values affecting the output. To address this, we assign a random value to [MASK] tokens, sampled from a Gaussian distribution $[MASK]_{i}^{(j)} \sim \text{N}(\mu_i^{(j)},\sigma_i^{(j)})$, where the mean and standard deviation are calculated from the standardization layer. This random approach makes it more challenging for the model to identify the positions of the [MASK] tokens, thereby encouraging a deeper understanding of the CSI sequence structure. For [PAD] token, even though we use a fixed value for [PAD], we replace them with [MASK] tokens before feeding the CSI sequence to the model. This is because there is no [PAD] in the CSI recovery phase. Even though the [PAD] can be ignored by the attention mechanism through the attention mask matrix, it still affects the output of the ARL, which works in the dimension of subcarrier. 
If [PAD] tokens only appear in the unsupervised training phase, it would cause a gap between inference and training, thereby influencing the model's performance. As a result, we also disable the \textcolor{black}{attention mask} in this phase.

For the unsupervised training phase, we follow a similar setting as RoBERTa \cite{RoBERTa} where some non-[PAD] tokens are randomly replaced with [MASK] tokens. In RoBERTa, the masking proportion is fixed at 15\%. However, in practical scenarios, the loss rate of CSI can vary, so we introduce a random masking proportion ranging from 15\% to 70\%, which is varied in each epoch.

Next, we train the model to recover the [MASK] tokens, referring to the model as the recoverer. To enhance the realism of the recovered CSI, we employ an additional discriminator model and train them using a GAN framework:
\begin{equation}
\begin{aligned}
& \underset{R}{\text{min}} \ \underset{D}{\text{max}} \ V(D,R) \\
& = \underset{R}{\text{min}} \ \underset{D}{\text{max}} \ E_c[log(D(c))] + E_c[log(1-D(R(c)))] \ ,
\label{GAN}
\end{aligned}
\end{equation}
where \textcolor{black}{$V$ is the target function}, $D$ and $R$ represent the discriminator and recoverer, respectively, and $c$ represents the original CSI input. The discriminator tries to distinguish the original CSI input and the recovered CSI sequence output by the recoverer. The recoverer, in turn, learns to generate outputs that are indistinguishable from real CSI.

Specifically, the loss function of the recoverer consists of four parts:
\begin{equation}
\begin{aligned}
& L_1=\text{MSE}(c,\hat{c}) \ , \\
& L_2=\text{MSE}(\mu,\hat{\mu}) \ , \\
& L_3=\text{MSE}(\sigma,\hat{\sigma}) \ ,\\
& L_4=\text{CrossEntropy}(D(\hat{c}),1) \ ,
\label{Loss}
\end{aligned}
\end{equation}
where $\hat{c}$ represents the output of the recoverer, and $\hat{\mu}$ and $\hat{\sigma}$ represent the mean and standard deviation of $\hat{c}$ in the time dimension, respectively. The $L_1$ loss is the traditional loss function used in BERT, as it measures the accuracy of the output. However, in CSI recovery, preserving the overall shape of the CSI is also important. Therefore, we utilize the $L_2$ and $L_3$ losses to consider the overall shape of the CSI by assessing its mean and standard deviation. Furthermore, the $L_4$ is the loss for the GAN training framework, which tries to make the discriminator mistake in distinguishing the recovered CSI. Additionally, we calculate all the loss functions focusing only on the [MASK] tokens again to ensure that the model prioritizes the recovery of the missing CSI. Moreover, the loss function of the discriminator is:
\begin{equation}
\begin{aligned}
L_{dis} = \text{CrossEntropy}(D(\hat{c}),0) + \text{CrossEntropy}(D(c),1) \ ,
\label{Loss_dis}
\end{aligned}
\end{equation}

To reduce overfitting, we also modify the training data selection strategy. Unlike CSI-BERT, which uses a sliding window, CSI-BERT2 randomly samples CSI sequences from the entire dataset during training. \textcolor{black}{The detailed pre-training algorithm is presented in Algorithm \ref{alg:pretraining}.}

\begin{algorithm}[htbp]
\caption{\textcolor{black}{Unsupervised Pre-training Process}}
\label{alg:pretraining}
\begin{algorithmic}[1]
\REQUIRE CSI dataset $\mathcal{D} = \{(c_i, t_i)\}_{i=1}^N$, pre-training epochs $E$, batch size $B$
\ENSURE Pre-trained CSI-BERT2 model $R$ (recoverer)
\STATE Initialize recoverer $R$ and discriminator $D$ with random weights
\STATE Preprocess data using Eq. (\ref{package loss}) and (\ref{timestamp inter}) to handle packet loss
\FOR{epoch $= 1$ to $E$}
    \FOR{each batch of CSI sequences $\{(c^{(j)}, t^{(j)})\}_{j=1}^B$}
        \STATE Standardize each CSI sequence using Eq. (\ref{standard})
        \STATE Randomly select masking ratio $r \sim \text{Uniform}(0.15, 0.70)$
        \STATE Replace [PAD] tokens with [MASK] tokens
        \STATE Randomly mask $r$ proportion of non-[PAD] tokens with [MASK] tokens
        \STATE Assign random values to [MASK] tokens: $[MASK]_i^{(j)} \sim \mathcal{N}(\mu_i^{(j)}, \sigma_i^{(j)})$
        \STATE $\hat{c} \gets R(c, t)$ \COMMENT{Forward pass through recoverer}
        \STATE Calculate recoverer loss $L_R$ using Eq. (\ref{Loss})
        \STATE Update $R$ by minimizing $L_R$ via backpropagation
        \STATE Calculate discriminator loss $L_{dis}$ using Eq. (\ref{Loss_dis})
        \STATE Update $D$ by minimizing $L_{dis}$ via backpropagation
    \ENDFOR
\ENDFOR
\RETURN Pre-trained model $R$
\end{algorithmic}
\end{algorithm}

\subsection{Fine-tuning and Inference Process in Downstream Tasks}
\subsubsection{CSI Prediction}

Encoder-based methods like BERT \cite{BERT} are often considered not suitable enough for generation or prediction tasks due to their bi-directional attention mechanisms. This is also influenced by their unsupervised training task, where MLM is used to train the model to recover the inner data, not the future data. However, decoder-based methods like GPT \cite{GPT} and encoder-decoder based methods like BART \cite{BART} also suffer from the problem of error accumulation \cite{transformer-time}. Since they generate data in an autoregressive manner, an error may also influence the further generation results. In contrast, encoder-based methods have the advantage that they can generate all data in one step.

To overcome this, CSI-BERT2 introduces a MPM training method. MPM has a similar format as MLM, but it only masks tokens at the end of the sequence instead of masking tokens in arbitrary random positions. Specifically, we randomly mask 15\% to 40\% of the tokens at the end of the sequence. Except for this, the other training strategy is exactly the same as the unsupervised training.

In the inference phase of the prediction task, we first use [MASK] to replace all [PAD] tokens, and then we fix the last 20\% of tokens in the input sequences as [MASK]. The model is then used to predict the last 20\% of tokens. \textcolor{black}{The detailed algorithm is similar to Algorithm \ref{alg:pretraining}, with the only difference being that the mask ratio $r$ is sampled from Uniform(0.15, 0.40), and the last $r$ tokens are replaced by [MASK].}

\subsubsection{CSI Classification}

For CSI classification tasks, we adopt the approach commonly used for pre-trained models, where we freeze the bottom layers of CSI-BERT2 and only train the top layers for task-specific adaptation. Different from CSI-BERT, in this phase, we use [MASK] tokens to replace not only all [PAD] tokens but also some other input tokens randomly, to imitate the situation of different packet loss rates. This improves generalization to real-world scenarios where packet loss rates may vary during deployment. \textcolor{black}{The detailed algorithm is presented in Algorithm \ref{alg:classification}.}

In the inference phase of the classification task, there is no need to add extra [MASK] tokens as in fine-tuning, but only to use [MASK] to replace [PAD]. Furthermore, thanks to our time embedding, CSI-BERT2 generalizes effectively to sequences with different sampling rates. Unlike traditional neural networks, which often fail under such variability, CSI-BERT2 maintains high performance even when the sampling rate of the testing data differs from that used in training.

\begin{algorithm}[htbp]
\caption{\textcolor{black}{CSI Classification Fine-tuning}}
\label{alg:classification}
\begin{algorithmic}[1]
\REQUIRE Pre-trained CSI-BERT2 model $R$, dataset $\mathcal{D} = \{(c_i, t_i, y_i)\}_{i=1}^N$, fine-tuning epochs $E_c$, batch size $B$
\ENSURE Fine-tuned CSI classification model $F \circ R$
\STATE Freeze bottom layers of CSI-BERT2 $R$, only train top classification head $F$
\STATE Preprocess data using Eq. (\ref{package loss}) and (\ref{timestamp inter})
\FOR{epoch $= 1$ to $E_c$}
    \FOR{each batch $\{(c^{(j)}, t^{(j)}, y^{(j)})\}_{j=1}^B$}
        \STATE Standardize each CSI sequence using Eq. (\ref{standard})
        \STATE Randomly select masking ratio $r \sim \text{Uniform}(0.15, 0.70)$
        \STATE Replace [PAD] tokens with [MASK] tokens
        \STATE Randomly mask $r$ proportion of non-[PAD] tokens with [MASK] tokens
        \STATE Assign random values to [MASK] tokens: $[MASK]_i^{(j)} \sim \mathcal{N}(\mu_i^{(j)}, \sigma_i^{(j)})$
        \STATE $\hat{y} \gets F(R(c, t))$ \COMMENT{Classification output}
        \STATE $L \gets \text{CrossEntropy}(y, \hat{y})$ \COMMENT{Classification loss}
        \STATE Update classification head $F$ by minimizing $L$ via backpropagation
    \ENDFOR
\ENDFOR
\RETURN Fine-tuned classification model $F \circ R$
\end{algorithmic}
\end{algorithm}

\section{Experiment} \label{Experiment}

\subsection{Experiment Configuration}
We illustrate our model configurations in Table \ref{configuration}. We trained our model on an NVIDIA RTX 3090 GPU. During training, we observed that CSI-BERT2 occupies approximately 2500MB of GPU memory. 

\begin{table}[htbp]
\caption{Model Configurations: This table provides a detailed overview of our CSI-BERT2 in the following experiments. In this context, `M' represents `million' and `K' represents `kilo', which will remain consistent in the subsequent tables.}
    \centering
        \begin{tabular}{|c|c|}
        \hline
        \textbf{Configuration} & \textbf{Our Setting} \\
        \hline 
        Input Length     & 100   \\
        \hline 
        Input Dimension     & 52   \\
        \hline
        Network Layers   & 6   \\
        \hline 
        Hidden Size   & 128    \\
        \hline 
        Inner Linear Size     & 512   \\
        \hline 
        Attn. Heads   & 8   \\
        \hline 
        Dropout Rate  & 0.1    \\
        \hline 
        Optimizer    & AdamW \\
        \hline 
        Learning Rate  & 0.0005  \\
        \hline 
        Batch Size   & 64  \\
        \hline 
        Total Number of Parameters  & 5.45M   \\
        \hline 
        \end{tabular}
\label{configuration}
\end{table}

\subsection{Dataset Description}
In our experiment, we utilize three real-collected Wi-Fi sensing datasets: the publicly available WiGesture dataset \cite{CSI-BERT}, the WiFall Dataset \cite{KNN-MMD}, and a novel dataset named WiCount, which is proposed along with this paper. All these datasets are collected using the ESP32-S3 device as RX and a home router as TX, and share the same data format, with a sample rate of 100Hz, 1 antenna, and 52 subcarriers. 
Additionally, considering the communication scenario, due to the limited availability of open-source public datasets, we generate a simulated dataset named CommPre using Sionna \cite{hoydis2022sionna}, which maintains the same format as other three sensing datasets.
 \textcolor{black}{In our experiment, we divide the data into 1-second samples, resulting in each sample having a length of 100 data points.We take the first 90\% of the samples as the training set and the remaining 10\% as the testing set. The details of each dataset are provided in the following subsections and Table \ref{tab:dataset}.}

\begin{table*}[htbp]
\caption{\textcolor{black}{Dataset Summarization}}
    \centering
        \begin{adjustbox}{width=\textwidth}
        \begin{tabular}{|c||c|c|c|}
        \hline
        \textbf{Dataset} & \textbf{Real Data} & \textbf{Time Duration (min)} & \textbf{Tasks} \\
        \hline 
        \textbf{WiGesture \cite{CSI-BERT} (dynamic part)} & $\checkmark$ & 48 & gesture recognition (6 classes), person identification (8 classes), CSI prediction\\
        \hline 
        \textbf{WiFall \cite{KNN-MMD}} & $\checkmark$ & 45 & action recognition (5 classes), fall detection (2 classes), CSI prediction \\
        \hline 
        \textbf{WiCount} & $\checkmark$ & 15 & people count estimation (4 classes), CSI prediction\\
        \hline 
        \textbf{CommPre} & $\times$ & 15 & CSI prediction \\
        \hline 
        \end{tabular}
        \end{adjustbox}
\label{tab:dataset}
\end{table*}

\subsubsection{WiGesture Dataset}
The WiGesture dataset\footnote{\textcolor{black}{\url{https://huggingface.co/datasets/RS2002/WiGesture}}} is a gesture recognition and person identification dataset, containing the CSI collected from 8 individuals performing 6 different gesture actions, including left-right, forward-backward, and up-down motions, clockwise circling, clapping, and waving. 

\subsubsection{WiFall Dataset}
The WiFall dataset\footnote{\textcolor{black}{\url{https://huggingface.co/datasets/RS2002/WiFall}}} is an action recognition, fall detection, and person identification dataset, containing the CSI collected from 10 individuals performing 5 different actions, including walking, jumping, sitting, standing up, and falling, which also consists of various types like forward fall, left fall, right fall, seated fall, and backward fall.

\subsubsection{WiCount Dataset}
\begin{figure}[htbp]
\centering 
\includegraphics[width=0.5\textwidth]{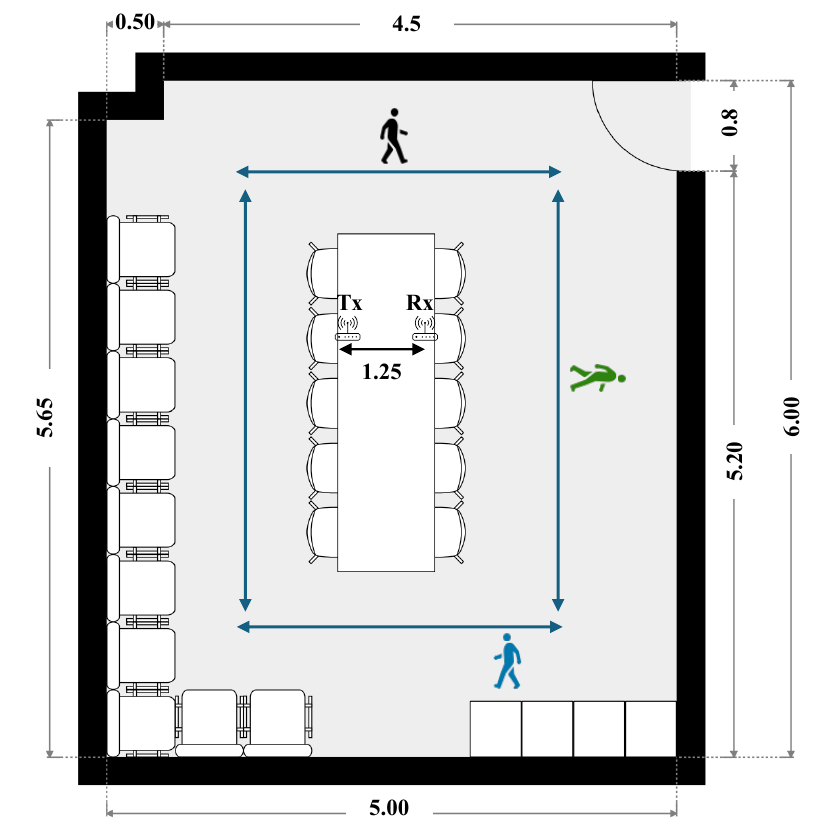}
\caption{Data collection environment of WiCount dataset, with measurements shown in meters.}
\label{env}
\end{figure}

\begin{figure*}
\centering 
\subfloat[Empty Room]{\includegraphics[width=0.25\textwidth]{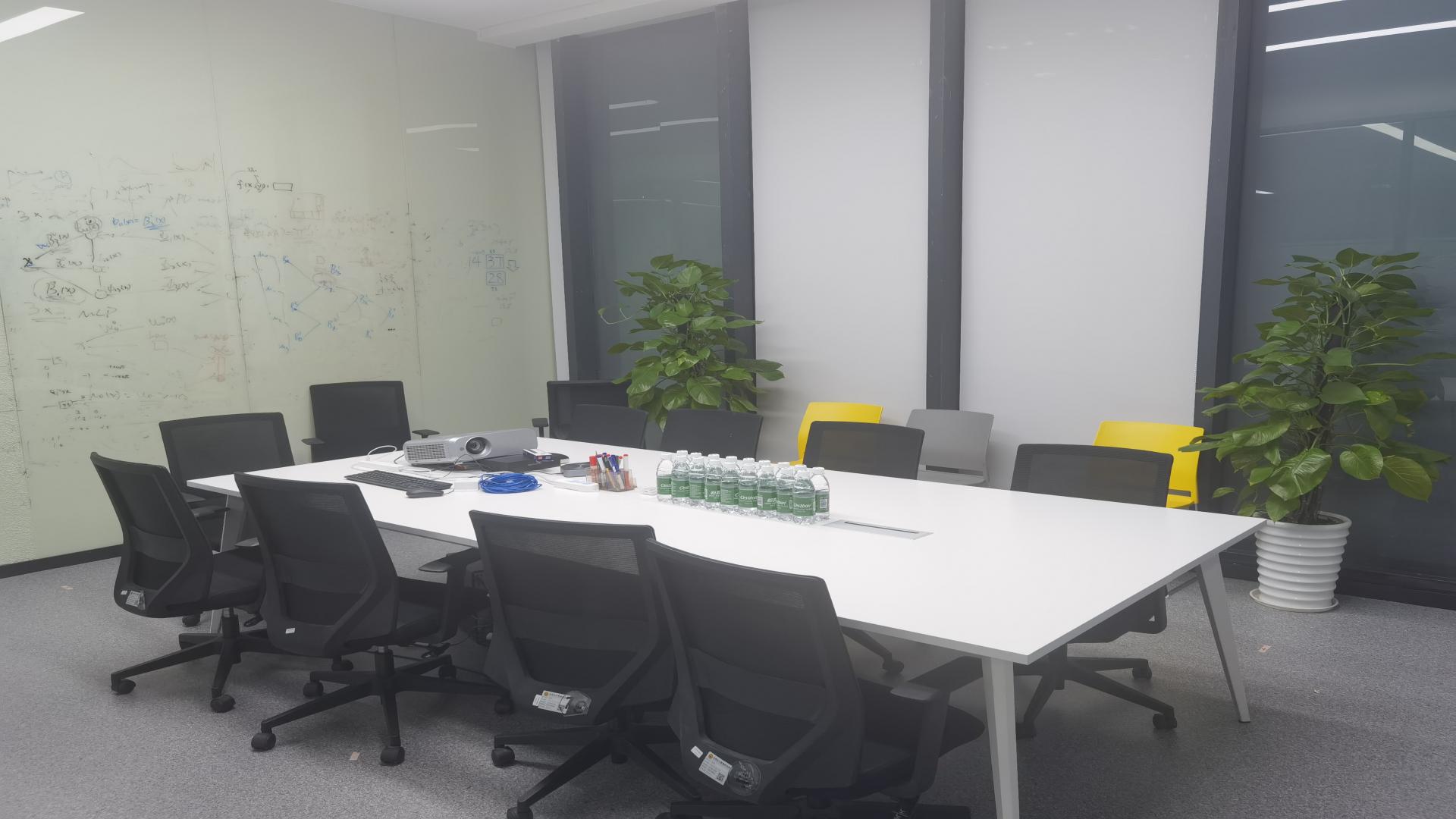}}
\subfloat[One Person]{\includegraphics[width=0.25\textwidth]{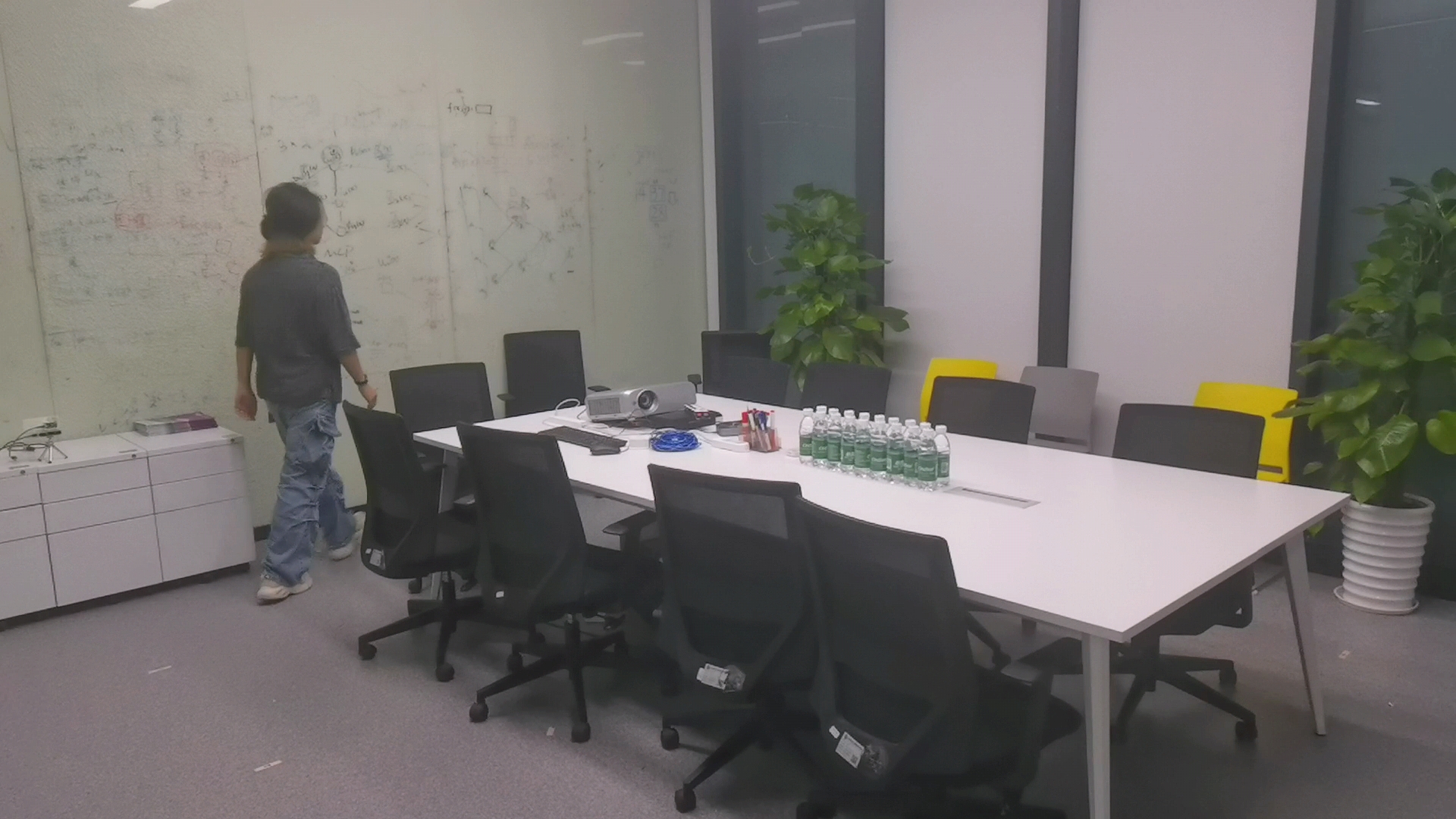}} 
\subfloat[Two People]{\includegraphics[width=0.25\textwidth]{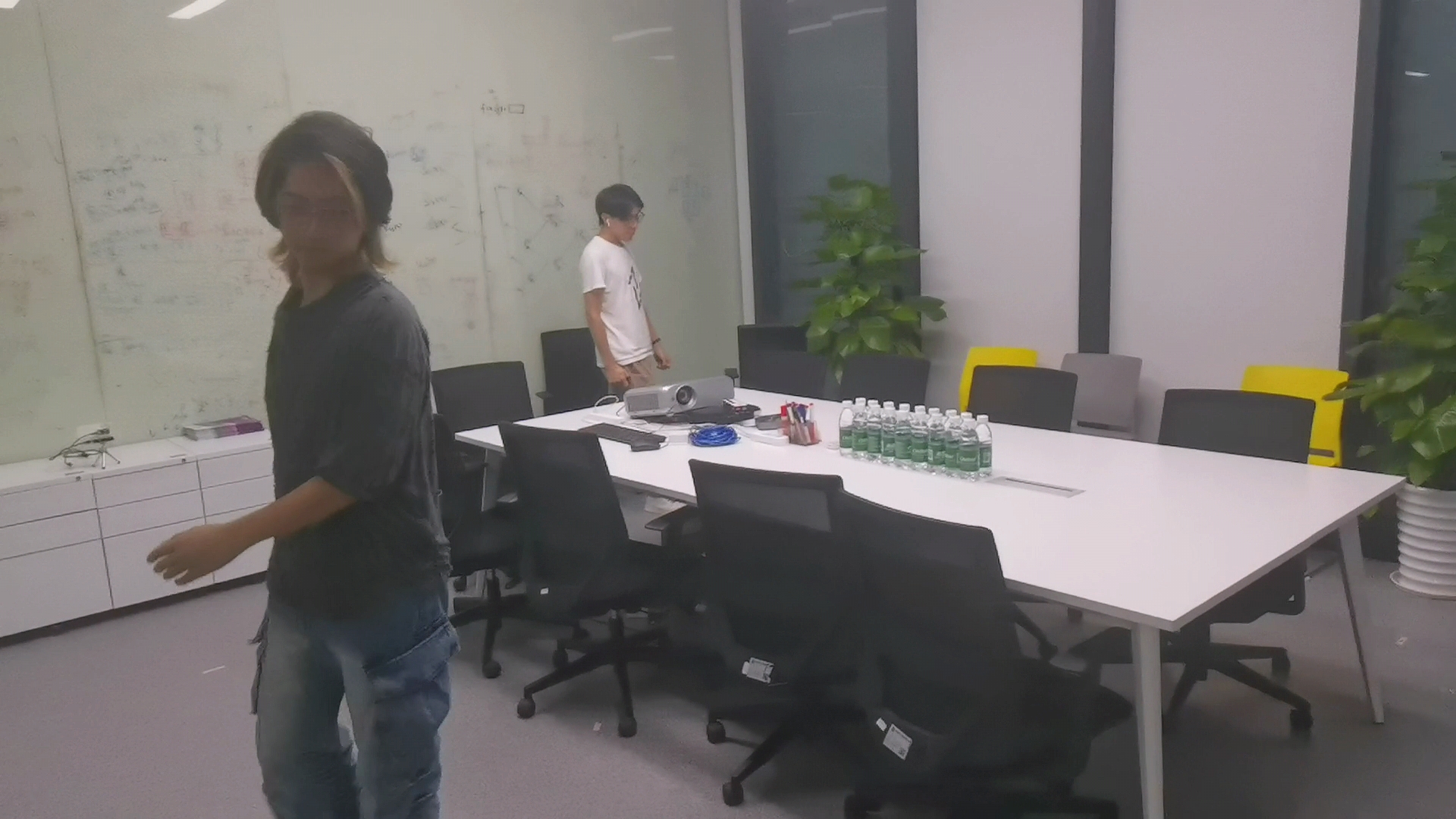}}
\subfloat[Three People]{\includegraphics[width=0.25\textwidth]{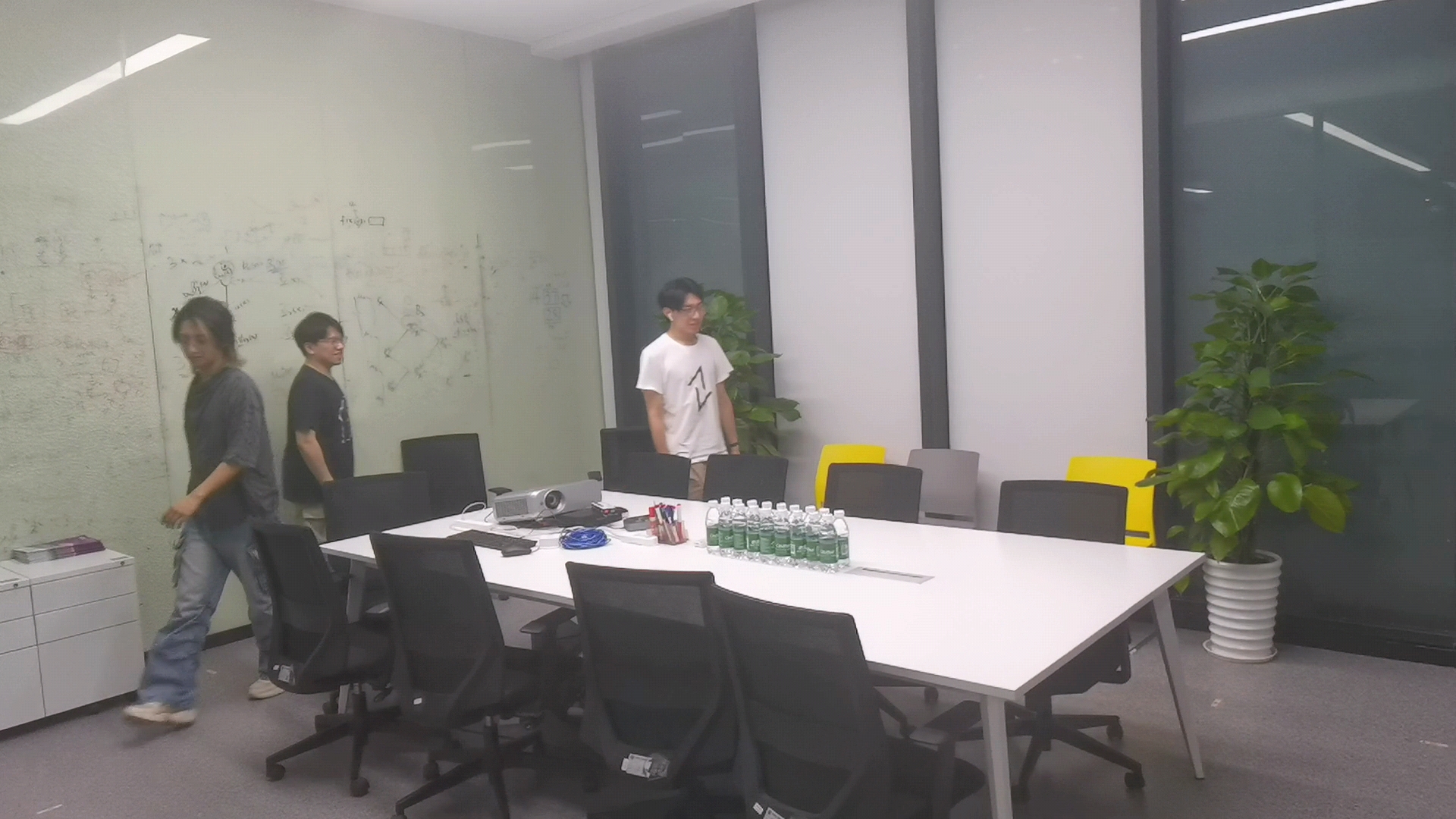}}
\caption{Sketch map of WiCount dataset.}
\label{dataset}
\end{figure*}

The WiCount dataset is a people number estimation dataset, collected in a small meeting room as shown in Fig. \ref{env}. The sketch map of the dataset is illustrated in Fig. \ref{dataset} that includes people number from 0 to 3. During data collection, individuals moved naturally, performing various activities such as walking, squatting, jumping, sitting, and moving in different directions.

\subsubsection{CommPre Dataset}
The CommPre dataset is generated using the Sionna RT\cite{sionna} platform with ray tracing techniques that capture multiple electromagnetic phenomena, including reflection, refraction, and diffraction. To realistically simulate challenging high-mobility environments, the positions of objects within the environment dynamically vary over time, with velocities ranging from 5 m/s to 20 m/s. Diverse mobility patterns are configured across different scenarios to represent various movement dynamics, and the corresponding multipath gains and phases are computed accordingly. Subsequently, the CSI is derived following Eq. (\ref{csi}), producing a CSI matrix of size 100 × 52 for each of the 892 simulated scenarios.

\subsection{Experiment Result}
\subsubsection{Unsupervised Training Result}

In this section, we first illustrate the performance of our model during the unsupervised training phase, as this directly influences the model's performance and convergence speed in downstream tasks. Following the approach taken by most signal generation studies \cite{he2023ai}, we evaluate unsupervised training performance using two metrics, including signal recovery error and the performance of the downstream model trained on the recovered signal.

The first dimension is the signal recovery error. To demonstrate that CSI-BERT2 effectively learns the CSI pattern, we compare the recovery performance of our method against that of CSI-BERT and other traditional interpolation methods. Specifically, we assess the recovery performance by randomly deleting 15\% of the packets and comparing the model's recovery results with the ground truth. We employ mean squared error (MSE), symmetric mean absolute percentage error (SMAPE), and mean absolute percentage error (MAPE) to quantify the error:
\begin{equation}
\begin{aligned}
& \text{MSE}(\hat{c},c) = \frac{1}{NM}\sum_{n=1}^{N}\sum_{m=1}^{M}(c^{(m)}_n - \hat{c}^{(m)}_n)^2 \ , \\
& \text{SMAPE}(\hat{c},c) = \frac{1}{NM}\sum_{n=1}^{N}\frac{|c^{(m)}_n - \hat{c}_n^{(m)}|}{(\frac{|c^{(m)}_n| + |\hat{c}^{(m)}_n|}{2}+\epsilon)} \ ,  \\
& \text{MAPE}(\hat{c},c) = \frac{1}{NM}\sum_{n=1}^{N}\frac{|c^{(m)}_n - \hat{c}^{(m)}_n|}{c^{(m)}_n+\epsilon} \ , 
\label{error}
\end{aligned}
\end{equation}
where $c$ and $\hat{c}$ represent ground truth and model output respectively, $N$ represents the number of packets, $M$ represents the number of subcarriers,$\epsilon$ is a small number used to avoid division by zero. The metrics are calculated only on the 15\% of the deleted CSI. Furthermore, to ensure fairness, we compute the error solely on the testing set, rather than on the entire dataset as conducted in CSI-BERT. We also compare the time required to recover the full dataset across different models, using only the CPU. Our results indicate that CSI-BERT2 achieves relatively low recovery error and recovery time across most datasets.

\begin{table*}[htbp]
\caption{Recovery Error: The bold value indicates the best result. The `Time (s)' represents the duration the models take to infer the entire testing set. This terminology will remain consistent across subsequent tables.}
    \centering
    \begin{adjustbox}{width=\textwidth}
        \begin{tabular}{|c||c|c|c|c||c|c|c|c||c|c|c|c|}
        \hline
        \textbf{Dataset} & \multicolumn{4}{c||}{\textbf{WiGesture} \cite{CSI-BERT}} & \multicolumn{4}{c||}{\textbf{WiFall}\cite{KNN-MMD}} & \multicolumn{4}{c|}{\textbf{WiCount}}\\
        \hline
        \textbf{\diagbox{Method}{Metric}} & \textbf{MSE} & \textbf{SMAPE} & \textbf{MAPE} & \textbf{Time (s)} & \textbf{MSE} & \textbf{SMAPE} & \textbf{MAPE} & \textbf{Time(s)} & \textbf{MSE} & \textbf{SMAPE} & \textbf{MAPE} & \textbf{Time (s)} \\
        \hline 
        CSI-BERT2 & \textbf{2.0800} & \textbf{0.1153} & \textbf{0.1217} & 5.53 & \textbf{4.1463} & \textbf{0.1240} & \textbf{0.1351} & 2.77 & 2.4531 & \textbf{0.1092} & 0.1189 & 1.56 \\
        \hline 
        CSI-BERT \cite{CSI-BERT} & 2.2438 & 0.1156 & 0.1244 & \textbf{1.84} & 4.4042 & 0.1271 & 0.1373 & \textbf{1.32} & \textbf{2.4471} & \textbf{0.1092} & 0.1185 & \textbf{0.67} \\
        \hline 
        Linear Interpolation & 2.8642 & 0.1266 & 0.1364 & 38.49 & 6.4420 & 0.1461 & 0.1571 & 2.81 & 2.6870 & 0.1099 & \textbf{0.1175} & 1.20 \\
        \hline 
        Ordinary Kringing & 3.5090 & 0.1390 & 0.1612 & 2709.43  & 4.6637 & 0.1319 & 0.1462 & 289.09 & 4.5964 & 0.1423 & 0.1684 & 109.10 \\
        \hline 
        Inverse Distance Weighted (IDW) & 2.4726 & 0.1187 & 0.1301 & 19.82 & 4.4251 & 0.1276 & 0.1409 & 2.45 & 3.4431 & 0.1268 & 0.1483 & 0.82 \\
        \hline 
        \end{tabular}
     \end{adjustbox}
\label{recovery}
\end{table*}

However, a low recovery error does not necessarily imply that the model has effectively learned the CSI pattern. For instance, the vanilla BERT model \cite{BERT} also achieves a relatively low recovery errors by producing similar outputs regardless of the input. Consequently, we adopt a common approach in generation studies by using the recovered data to train other models. If these models perform well on the generated data, it suggests that the recovered data is likely to be close to reality. Following the methodology of CSI-BERT \cite{CSI-BERT}, we provide two methods for CSI recovery, named ``recover" and ``replace":
\color{black}
\begin{equation}
\begin{aligned}
& CSI^{replace}=\hat{c} \ , \\
& CSI^{recover}=(1-IsPad) \cdot c + IsPad  \cdot \hat{c} \ , \\
& IsPad=\textbf{1}\{c==[PAD]\} \ , 
\label{recover}
\end{aligned}
\end{equation}
where $\textbf{1}\{ \cdot \}$ is the indicator function that returns 1 when the condition is true and 0 otherwise, $IsPad$ indicates whether the position of the original CSI $c$ corresponds to the [PAD] token.
\color{black}

Subsequently, we utilize the data recovered by different methods to train various networks and evaluate their accuracy across multiple tasks, including gesture recognition, person identification, action recognition, fall detection, and crowd estimation. The models for comparision include traditional neural networks such as MLP, recently proposed architectures like LSTM \cite{LSTM}, and Wi-Fi sensing models such as WiGRUNT \cite{WiGRUNT}. The results are summarized in Table \ref{sensing}. The findings clearly indicate that using CSI-BERT2 for data recovery yields the most significant improvements compared to other methods, suggesting that the data recovered by CSI-BERT2 is the closest to reality.



\begin{table*}[htbp]
\caption{CSI Classification Performance: The number below each model name represents the number of parameters. The bold value indicates the best result in each column, and the underlined value indicates the best result in each row within each task.}
    \centering
        \begin{adjustbox}{width=\textwidth}
        \begin{tabular}{|c||c|c|c|c|c|c|c|c||c|}
        \hline
        \textbf{Task} & \multicolumn{9}{c|}{\textbf{Gesture Recognition (WiGesture Dataset \cite{CSI-BERT})}}\\
        \hline
        \multirow{2}{*}{\textbf{\diagbox{Data}{Model}}} & \textbf{MLP} & \textbf{CNN} & \textbf{RNN \cite{RNN}} & \textbf{LSTM \cite{LSTM}} &  \textbf{Chen et al. \cite{fall}}  & \textbf{WiGRUNT \cite{WiGRUNT}} & \textbf{CSI-BERT \cite{CSI-BERT}} & \textbf{CSI-BERT2} & \multirow{2}{*}{\textbf{Average}} \\
        &  337K & 23K & 33K & 133K & 11M & 11M & 2M & 5M &  \\
        \hline 
        Original Data & 66.93\% & 55.72\% & 39.56\% & 11.97\% & 70.31\% & 48.73\% & 76.91\% & \underline{\textbf{99.48\%}} & 58.70\% \\
        \hline
         CSI-BERT2 recover & 72.88\% & 57.27\% & 54.34\% & 48.35\% & \textbf{\underline{92.96\%}} & \textbf{78.97\%} & \textbf{92.18\%} & 89.06\% & 73.25\% \\
         \hline 
         CSI-BERT2 replace & 73.68\% & \textbf{62.80\%} & 55.48\% & 40.79\% & 91.92\% & 74.99\% & 81.51\% & \underline{91.95\%} & 71.63\% \\
        \hline 
         CSI-BERT  recover & 74.23\% & 59.39\% & 48.96\% & 22.92\% & 92.57\% & 71.87\% & 71.87\% & \underline{92.70\%} & 66.81\% \\
         \hline 
         CSI-BERT replace & \textbf{86.90\%} & 61.51\% & \textbf{58.80\%} & 52.36\% & 84.52\% & 78.84\% & 79.54\% & \underline{91.41\%} & \textbf{74.24\%} \\
        \hline 
         Linear Interpolation & 72.91\% & 58.35\%  & 45.32\% & 49.09\% & 80.75\% & 74.91\% & 74.55\% & \underline{88.25\%} & 68.01\% \\
        \hline 
        Ordinary Kringing & 65.62\% & 57.55\% & 53.64\% & \textbf{50.00\%} & \underline{88.71\%} & 69.99\% & 74.27\% & 85.93\% & 68.21\% \\
        \hline 
        IDW & 40.17\% & 56.77\%  & 48.70\% & 46.88\%  & 80.32\% & 71.06\% & 67.22\% &  \underline{88.28\%} & 62.42\% \\
        \hline \hline 
        \textbf{Task} & \multicolumn{9}{c|}{\textbf{Person Identification (WiGesture Dataset \cite{CSI-BERT})}}\\
        \hline
        \textbf{Model} & \textbf{MLP} & \textbf{CNN} & \textbf{RNN \cite{RNN}} & \textbf{LSTM \cite{LSTM}} & \textbf{Chen et al. \cite{fall}}  & \textbf{WiGRUNT \cite{WiGRUNT}} & \textbf{CSI-BERT \cite{CSI-BERT}} & \textbf{CSI-BERT2} & \textbf{Average} \\
        \hline
        Original Data & 71.34\% & 71.14\% & 66.39\% & 21.09\%  & 83.76\% & 72.07\% & 93.94\% & \underline{\textbf{99.73\%}} & 72.43\% \\
        \hline
         CSI-BERT2 recover & 95.57\% & \textbf{85.54\%} & 84.60\% & 27.98\% & 93.20\% & 81.73\% & \textbf{97.92\%} & \underline{\textbf{99.73\%}} & 83.28\% \\
         \hline 
         CSI-BERT2 replace & 95.05\% & 83.07\% & 84.68\% & \textbf{54.13\%} & 95.33\% & 83.84\% & \underline{96.35\%} & 94.79\% & \textbf{85.91\%} \\
        \hline 
         CSI-BERT  recover & \textbf{97.13\%} & 80.60\% & 80.51\% & 35.18\%  & 94.30\% & \textbf{84.67\%} & 95.05\% &\underline{\textbf{99.73\%}} & 83.39\% \\
         \hline 
         CSI-BERT replace & 97.65\% & 79.18\% & \textbf{89.24\%} & 24.22\% & \textbf{97.39\%} & 77.77\% & 95.83\% & \underline{99.47\%} & 82.59\% \\
        \hline 
         Linear Interpolation & 81.84\% & 70.88\% & 84.45\% & 26.83\% & 86.75\% & 70.28\% & \textbf{\underline{97.92\%}} & 91.67\% & 76.33\% \\
        \hline 
        Ordinary Kringing & 94.76\% & 85.38\% & 86.42\% & 21.61\% & 97.32\% & 80.84\% & 95.83\% & \underline{99.03\%} & 82.64\% \\
        \hline 
        IDW & 83.22\%  & 74.56\% & 88.54\% & 33.91\% & 94.27\% & 80.70\% & 95.20\% &  \underline{99.47\%} & 81.23\% \\
        \hline \hline 
        \textbf{Task} & \multicolumn{9}{c|}{\textbf{Action Recognition (WiFall Dataset \cite{KNN-MMD})}}\\
        \hline
        \textbf{Model} & \textbf{MLP} & \textbf{CNN} & \textbf{RNN \cite{RNN}} & \textbf{LSTM \cite{LSTM}} & \textbf{Chen et al. \cite{fall}}  & \textbf{WiGRUNT \cite{WiGRUNT}} & \textbf{CSI-BERT \cite{CSI-BERT}} & \textbf{CSI-BERT2} & \textbf{Average} \\
        \hline
        Original Data & 47.48\% & 56.27\% &  58.61\% & 52.10\% & 51.38\% & 34.44\% & \textbf{82.43\%} & \textbf{\underline{88.59\%}} & 58.91\% \\
        \hline
         CSI-BERT2 recover & 64.97\% & \textbf{67.18\%} & 68.48\% & 62.63\% & 71.70\% & 70.96\% & 67.63\% & \underline{72.16\%} & 68.21\% \\
         \hline 
         CSI-BERT2 replace & 69.01\% & 66.27\% & \textbf{70.18\%} & 61.99\% & \underline{73.96\%} & 69.72\% & 66.77\% & 72.70\% & \textbf{68.82\%} \\
        \hline 
         CSI-BERT  recover & 66.40\% & 54.94\% & 68.48\% & 61.79\% & 69.66\% & 70.94\% & 67.36\% & \underline{73.69\%} & 66.65\% \\
         \hline 
         CSI-BERT replace & \textbf{73.05\%} & 54.97\% & 66.79\% & \textbf{66.73\%} & 72.01\% & 67.44\% & 66.61\% & \underline{73.67\%} & 67.65\% \\
        \hline 
         Linear Interpolation & 67.44\% & 64.32\% & 67.31\% & 59.78\% & \textbf{\underline{74.22\%}} & 70.57\% & 64.37\% & 74.19\% & 67.77\% \\
        \hline 
        Ordinary Kringing & 67.96\% & 65.52\% & 64.44\% & 63.88\% & 70.92\% & 62.41\% & 67.36\% & \underline{71.77\%} & 66.78\% \\
        \hline 
        IDW & 70.31\% & 67.08\% & 69.79\% & 62.32\% & 71.09\% & \textbf{72.39\%} & 67.22\% & 70.21\% & 68.80\% \\
        \hline \hline 
        \textbf{Task} & \multicolumn{9}{c|}{\textbf{Fall Detection (WiFall Dataset \cite{KNN-MMD})}}\\
        \hline
        \textbf{Model} & \textbf{MLP} & \textbf{CNN} & \textbf{RNN \cite{RNN}} & \textbf{LSTM \cite{LSTM}} & \textbf{Chen et al. \cite{fall}}  & \textbf{WiGRUNT \cite{WiGRUNT}} & \textbf{CSI-BERT \cite{CSI-BERT}} & \textbf{CSI-BERT2} & \textbf{Average} \\
        \hline
        Original Data & 78.34\% &  52.99\% & 82.29\% & 80.35\% & 78.52\% & 73.69\% & \textbf{93.28\%} & \textbf{\underline{94.79\%}} & 79.28\% \\
        \hline
         CSI-BERT2 recover &  80.79\% & \textbf{75.95\%} & \textbf{86.58\%} & \textbf{86.72\%} & \textbf{82.42\%} & 80.90\% & 82.25\% & \underline{86.97\%} & \textbf{82.82\%} \\
         \hline 
         CSI-BERT2 replace & 79.82\%  & 74.89\% & 83.07\% & \underline{86.31\%} & 82.16\% & 78.65\% & 80.62\% & 85.38\% & 81.36\% \\
        \hline 
         CSI-BERT  recover & 80.98\%  & 75.27\% & 84.37\% & 80.41\% & 81.38\% & \textbf{83.07\%} & 81.32\% & \underline{84.92\%} & 81.46\% \\
         \hline 
         CSI-BERT replace &  80.21\% & 74.94\% & 83.46\% & 84.37\% & 82.33\% & 80.79\% & 83.33\% & \underline{85.72\%} & 81.89\% \\
        \hline 
         Linear Interpolation & 81.78\%  & 75.78\% & 84.50\% & 84.33\% & 78.51\% & 78.77\% & 81.35\% & \underline{84.39\%} & 81.17\% \\
        \hline 
        Ordinary Kringing & 81.64\% &  75.78\% & 80.98\% & 82.29\% & 82.00\% & 79.03\% & 82.31\% & \underline{84.49\%} & 81.07\% \\
        \hline 
        IDW & \textbf{82.55\%} & 54.94\% & 83.59\% & 80.59\% & 78.21\% & 80.72\% & 81.72\% & 84.06\% & 78.29\% \\
        \hline \hline 
        \textbf{Task} & \multicolumn{9}{c|}{\textbf{People Number Estimation (WiCount Dataset)}}\\
        \hline
        \textbf{Model} & \textbf{MLP} & \textbf{CNN} & \textbf{RNN \cite{RNN}} & \textbf{LSTM \cite{LSTM}} & \textbf{Chen et al. \cite{fall}}  & \textbf{WiGRUNT \cite{WiGRUNT}} & \textbf{CSI-BERT \cite{CSI-BERT}} & \textbf{CSI-BERT2} & \textbf{Average} \\
        \hline
        Original Data & 56.77\% & 69.68\% & 80.93\% & 80.72\% & 48.33\% & 49.53\% & \textbf{89.67\%} & \textbf{\underline{94.32\%}} & 71.24\% \\
        \hline
         CSI-BERT2 recover & 87.29\% & 78.75\% & 83.98\%  & 81.51\% & 83.32\% & \textbf{\underline{85.42\%}} & 84.06\% & \underline{91.32\%} & 84.45\% \\
         \hline 
         CSI-BERT2 replace & 85.62\% & 78.49\% & \textbf{88.12\%} & 86.97\% & 81.41\% & 82.34\% & 81.51\% & \underline{92.76\%} & \textbf{84.65\%} \\
        \hline 
         CSI-BERT  recover & \textbf{88.98\%} & 80.83\% & 86.20\% & 82.39\% & 82.22\% & 82.70\% & 79.04\% & \underline{92.70\%} & 84.38\% \\
         \hline 
         CSI-BERT replace & 81.61\% & 72.60\% & 85.67\% & 84.95\% & \textbf{85.62\%} & 82.75\% & 81.61\% & \underline{92.86\%} & 83.46\% \\
        \hline 
         Linear Interpolation & 76.51\% & 77.73\% & 85.52\%  & 82.40\%  & 80.17\% & 83.07\% & 86.51\% & \underline{88.64\%} & 82.57\% \\
        \hline 
        Ordinary Kringing & 87.29\% & 50.52\% & 84.84\% & 85.05\% & 82.97\% & 76.72\% & 85.72\% & \underline{91.90\%} & 80.63\% \\
        \hline 
        IDW & 80.72\% & \textbf{82.29\%} & 84.17\% & \textbf{87.00\%} & 82.10\% & 81.72\% & 85.62\% & \underline{88.54\%} & 84.02\% \\
        \hline            
        \end{tabular}
        \end{adjustbox}
\label{sensing}
\end{table*}

\subsubsection{CSI Prediction Task}
Given the scarcity of real CSI datasets in wireless communications, we adopt the widely used approach of leveraging wireless sensing datasets for performance evaluation.
For the prediction task, we provide the first 0.8 seconds of CSI in each sample and train the model to predict the subsequent 0.2 seconds of CSI. We compare the performance of our CSI-BERT2 model against several other CSI prediction models, including popular time series models like LSTM \cite{LSTM}, RNN \cite{RNN}, GRU \cite{GRU}, and Mamba \cite{Mamba}, which are also widely used in CSI prediction \textcolor{black}{\cite{LSTM_CSI1,LSTM_CSI2,LSTM_CSI3,LSTM_CSI4,RNN_CSI,GRU_CSI}}. We also compare our model with some CSI prediction networks like OCEAN \cite{OCEAN}, CV-3DCNN \cite{3DCNN}, and ConvLSTM \cite{ConvLSTM}. The prediction error and time cost (using GPU) are shown in Table \ref{prediction}. It can be seen that our CSI-BERT2 model significantly outperforms all other models. Particularly in the WiCount dataset, where the task is more challenging, as each gesture or action has a relatively irregular pattern, other methods exhibit a substantial decrease in performance, while CSI-BERT2 is influenced only minimally. \textcolor{black}{Although the inference time of CSI-BERT2 is longer than that of the other models, its total inference time for the entire testing set across different datasets remains under 0.5 seconds, with a sample volume of 500 for each dataset using a sliding window. Consequently, the inference time for a single sample is within 1 ms, which is acceptable in practice. Additionally, we provide the Floating Point Operations (FLOPs) of different methods for a single input in the table. Our CSI-BERT2 model requires higher FLOPs, but it only reaches around 0.5G, making it suitable for real-time applications in edge devices \cite{wang2025revisiting,wang2025ultra}.}

\begin{table*}[htbp]
\caption{\color{black}{CSI Prediction Error}}
    \centering
        \begin{tabular}{|c|c|c||c|c|c|c||c|c|c|c|}
        \hline
         \multicolumn{3}{|c||}{\textbf{Configuration}} & \multicolumn{4}{c||}{\textbf{CommPre}} & \multicolumn{4}{c|}{\textbf{WiGesture} \cite{CSI-BERT}} \\
        \hline
        \textbf{Model} & \textbf{Size} & \textbf{GFLOPs} & \textbf{MSE ($\times 10^{-8}$)} & \textbf{SMAPE} & \textbf{MAPE} & \textbf{Time(s)} & \textbf{MSE} & \textbf{SMAPE} & \textbf{MAPE} & \textbf{Time(s)} \\
        \hline 
        CSI-BERT2 & 5M & 0.5159 & \textbf{1.5689} & \textbf{0.2959} & \textbf{0.2319}  & 0.37 & \textbf{3.2942} & \textbf{0.1583} & \textbf{0.1349} & 0.46 \\
        \hline 
        LSTM \cite{LSTM,LSTM_CSI1,LSTM_CSI2,LSTM_CSI3,LSTM_CSI4} & 133K & 0.0134  & 5.1107 & 0.4849 & 0.5722 &  \textbf{0.03} & 12.3254 & 0.2397 & 0.3967 & 0.05  \\
        \hline 
        RNN \cite{RNN,RNN_CSI} & 33K & \textbf{0.0033} & 4.6164 & 0.4607  & 0.5385 & \textbf{0.03} & 19.4708 & 0.2877 & 0.4063 & \textbf{0.04}  \\
        \hline 
        GRU \cite{GRU,GRU_CSI} & 100K & 0.0100 & 5.4009 & 0.5128 & 0.6367 & 0.04 & 19.7180 & 0.2922 & 0.4243 & \textbf{0.04}  \\
        \hline 
        Mamba \cite{Mamba} & 5M & 0.5062  &  5.1286 & 0.4430 & 0.4838 & 0.23 & 12.3281 & 0.2392 & 0.3277 & 0.24 \\
        \hline 
        OCEAN \cite{OCEAN} & 126K & 0.0126 & 5.6013 & 0.4863 & 0.5576 & 0.05 & 19.6257 & 0.2925 & 0.4231 & 0.05   \\
        \hline
        CV-3DCNN \cite{3DCNN} & \textbf{19K} & 0.1033  & 4.5276 & 0.4794 & 0.6810 & 0.04 & 11.3017 & 0.2267 & 0.3044 & \textbf{0.04} \\
        \hline 
        ConvLSTM  \cite{ConvLSTM} & 152K & 0.1136 & 5.0015 & 0.4898 & 0.6031 & 0.04 & 19.7038 & 0.2921 & 0.4242 & \textbf{0.04} \\
        \hline \hline
         \multicolumn{3}{|c||}{\textbf{Configuration}} & \multicolumn{4}{c||}{\textbf{WiFall}\cite{KNN-MMD}} & \multicolumn{4}{c|}{\textbf{WiCount}}\\
        \hline
        \textbf{Method} & \textbf{Size} & \textbf{GFLOPs}  & \textbf{MSE} & \textbf{SMAPE} & \textbf{MAPE} & \textbf{Time(s)} & \textbf{MSE} & \textbf{SMAPE} & \textbf{MAPE} & \textbf{Time(s)} \\
        \hline 
        CSI-BERT2 & 5M & 0.5159 & \textbf{4.8598} & \textbf{0.1471}  & \textbf{0.1347} & 0.49 & \textbf{5.3401} & \textbf{0.1726} & \textbf{0.1590} & 0.46 \\
        \hline 
        LSTM  \cite{LSTM,LSTM_CSI1,LSTM_CSI2,LSTM_CSI3,LSTM_CSI4} & 133K & 0.0134 & 7.1495 & 0.1624 & 0.1882 & 0.04 & 32.3377 & 0.2547 & 0.3528 & 0.05 \\
        \hline 
        RNN  \cite{RNN,RNN_CSI} & 33K & \textbf{0.0033} & 16.9083 & 0.2424 & 0.2988 & 0.04 & 32.3670 & 0.2548 & 0.3534 & 0.03 \\
        \hline 
        GRU  \cite{GRU,GRU_CSI} & 100K & 0.0100 & 16.5353 & 0.2395 & 0.2963 & 0.07 & 39.8108 & 0.2541 & 0.3556 & \textbf{0.02} \\
        \hline 
        Mamba \cite{Mamba} & 5M & 0.5062  & 6.4666 & 0.1532 & 0.1756 & 0.12 & 39.9170 & 0.2566 & 0.3524 & 0.11 \\
        \hline 
        OCEAN  \cite{OCEAN} & 126K & 0.0126 & 16.8825 & 0.2423 & 0.2978 & \textbf{0.03} & 39.7917 & 0.2542 & 0.3548 & \textbf{0.02}  \\
        \hline
        CV-3DCNN  \cite{3DCNN} & \textbf{19K} & 0.1033  & 8.2616 & 0.1713 & 0.1981 & \textbf{0.03} & 42.2662 & 0.2631 & 0.3560 & \textbf{0.02} \\
        \hline 
        ConvLSTM \cite{ConvLSTM} & 152K & 0.1136  & 16.8935 & 0.2429 & 0.2983 & \textbf{0.03} & 39.7709 & 0.2537 & 0.3552 & \textbf{0.02}  \\
        \hline 
        \end{tabular}
\label{prediction}
\end{table*}

\subsubsection{CSI Classification Task} \label{CSI Sensing Task}
The robust and powerful capabilities of our CSI-BERT2 model in CSI classification tasks are demonstrated in Table \ref{sensing}, where it achieves the best performance across all scenarios. Notably, the original CSI-BERT fails to outperform models such as Chen et al. \cite{fall} and WiGrumt \cite{WiGRUNT} in many scenarios. However, our CSI-BERT2 shows substantial improvement due to the enhanced model structure, thanks to the ARL capturing the relationships and importance among subcarriers, and the MLP-based time embedding layer that encodes time information more effectively.
Moreover, aided by the time embedding mechanism, CSI-BERT2 can be effectively applied in scenarios where the training data includes samples with varying sampling rates, or where the sampling rate of the testing set differs from that of the training set.
This is a practical feature, especially in federated learning scenarios, where we cannot guarantee that all clients have the same sampling rate. 
\textcolor{black}{In Table \ref{sample rate}, we compare the performance of different models where the training and testing sets both have 100Hz and 50Hz CSI, as well as where the training set only has 100Hz CSI but the testing set has 50Hz CSI.}
For fairness, we provide the CSI processed by linear interpolation for the other non-CSI-BERT models. It can be seen that other models are significantly influenced by the different sampling rates, with some even failing to function properly in particular scenarios. In contrast, CSI-BERT2 is almost unaffected by any impact, thanks to the help of time embedding.

What's more, resampling methods could also be a potential way to address such scenarios. However, on one hand, the resampling operation may introduce a relatively large delay. On the other hand, these resampling methods were not designed specifically for CSI, which means they may not be able to capture the inherent relationships within the CSI as traditional interpolation methods. Therefore, the resampling error could also potentially harm the model's performance.

\begin{table*}[htbp]
\caption{CSI Classification Performance under Different Sampling Rate}
    \centering
        \begin{adjustbox}{width=\textwidth}
        \begin{tabular}{|c||c|c|c|c|c|c|c|c|}
        \hline
        \textbf{Data} & \multicolumn{8}{c|}{\textbf{Training Set: 100Hz+50Hz; Testing Set: 100Hz+50Hz}}\\
        \hline
        \textbf{Model} & \textbf{MLP} & \textbf{CNN} & \textbf{RNN \cite{RNN}} & \textbf{LSTM \cite{LSTM}}  & \textbf{Chen et al. \cite{fall}}  & \textbf{WiGRUNT \cite{WiGRUNT}} & \textbf{CSI-BERT \cite{CSI-BERT}} & \textbf{CSI-BERT2} \\
        \hline
        Gesture Recognition & 16.51\%  & 14.54\% & 15.13\% & 17.37\% & 16.59\% & 74.47\% & 64.61\%  & \textbf{97.04\%} \\
        \hline
         Person Identification & 13.47\%  & 15.52\% & 13.41\% & 13.47\% & 13.72\% & 81.25\% & 70.83\% & \textbf{99.54\%} \\
         \hline 
         Action Recognition & 70.97\% & 67.67\%  & 70.02\% & 59.63\% & 75.18\% & 66.78\% & 78.18\% & \textbf{88.35\%} \\
        \hline 
         Fall Detection & 81.10\%  & 75.80\% &  83.92\% & 84.83\% & 85.98\% & 80.72\% & 92.98\% & \textbf{93.64\%} \\
         \hline 
         People Number Estimation & 84.03\% & 46.82\% & 81.25\% & 82.00\% & 80.09\% & 76.13\% & 86.93\% & \textbf{92.54\%} \\
        \hline \hline     
        \textbf{Data} & \multicolumn{8}{c|}{\textbf{Training Set: 100Hz; Testing Set: 50Hz}}\\
        \hline
        \textbf{Model} & \textbf{MLP} & \textbf{CNN} & \textbf{RNN \cite{RNN}} & \textbf{LSTM \cite{LSTM}} & \textbf{Chen et al. \cite{fall}}  & \textbf{WiGRUNT \cite{WiGRUNT}} & \textbf{CSI-BERT \cite{CSI-BERT}} & \textbf{CSI-BERT2} \\
        \hline
        Gesture Recognition & 69.79\% & 20.38\% & 36.25\%  & 27.15\% & 74.89\% & 71.37\% & 79.96\% & \textbf{97.81\%} \\
        \hline
         Person Identification & 87.29\% & 11.85\%  & 82.29\% & 22.32\% & 87.22\% & 85.24\% & 94.44\% & \textbf{99.38\%} \\
         \hline 
         Action Recognition & 68.97\% & 51.96\% & 68.07\%  & 60.26\% & 76.56\% & 73.21\% & 84.56\% & \textbf{88.53\%} \\
        \hline 
         Fall Detection & 80.83\% & 76.13\% & 80.17\% & 84.38\% & 78.61\% & 77.08\% & 94.19\% & \textbf{94.32\%} \\
         \hline 
         People Number Estimation & 82.22\%  & 77.03\% & 84.43\% & 42.22\% & 68.89\% & 71.11\% & 89.10\% & \textbf{94.77\%} \\
        \hline 
        \end{tabular}
        \end{adjustbox}
\label{sample rate}
\end{table*}

\subsection{Ablation Study}
\subsubsection{Effect of Unsupervised Pre-Training}

\begin{figure*}[htbp]
\centering 
\includegraphics[width=\textwidth]{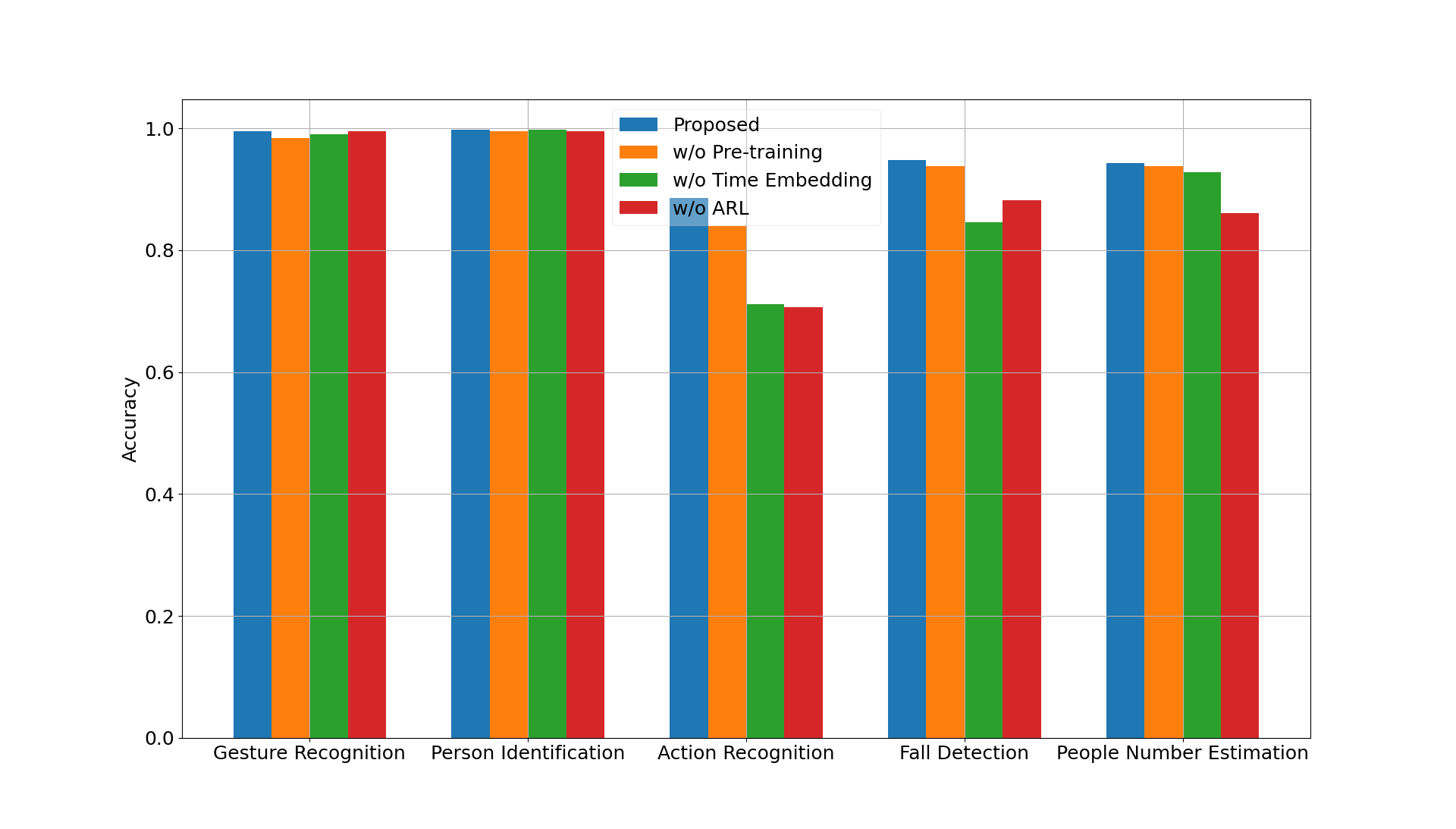}
\caption{\textcolor{black}{CSI-BERT2 performance in CSI classification task}}
\label{fig:pretrain}
\end{figure*}

To illustrate the benefits of our unsupervised training mechanism for downstream tasks, we compare the model's performance with and without unsupervised training, as shown in Fig. \ref{fig:pretrain} and Table \ref{tab:pretrain}. 
While unsupervised training contributes to performance gains in the CSI classification task, the improvement remains relatively modest.
However, in the CSI prediction task, unsupervised training proves to be highly beneficial, this is because the prediction task is more closely related to the mask recovery task.

\begin{table*}[htbp]
\caption{\color{black}{CSI-BERT2 Performance in CSI Prediction Task}}
    \centering
        \begin{tabular}{|c||c|c|c||c|c|c|}
        \hline
        \multirow{2}{*}{\textbf{Dataset}} & \multicolumn{3}{c||}{\textbf{Proposed}} & \multicolumn{3}{c|}{\textbf{w/o Unsupervised Training}}\\
        \cline{2-7}
         & \textbf{MSE} & \textbf{SMAPE} & \textbf{MAPE} & \textbf{MSE} & \textbf{SMAPE} & \textbf{MAPE} \\
        \hline 
        CommPre  & \textbf{1.5689 $\times 10^{-8}$} & \textbf{0.2959} & 0.2319   & 2.0080 $\times 10^{-8}$ & 0.3198 & 0.2477 \\
        \hline
        WiGesture \cite{CSI-BERT}  & \textbf{3.2942} & \textbf{0.1583} & \textbf{0.1349}  & 5.3054 & 0.1962 & 0.1657 \\
        \hline
        WiFall \cite{KNN-MMD}  & \textbf{4.8598} & \textbf{0.1471} & \textbf{0.1347}  &  5.0957 & 0.1595 & 0.1413 \\
        \hline
        WiCount  & \textbf{5.4301} & \textbf{0.1726} & 0.1590 & 6.6868 & 0.2019 & 0.1659 \\
        \hline \hline
        \multirow{2}{*}{\textbf{Dataset}} & \multicolumn{3}{c||}{\textbf{w/ Time Embedding}} & \multicolumn{3}{c|}{\textbf{w/o ARL}}\\
        \cline{2-7}
         & \textbf{MSE} & \textbf{SMAPE} & \textbf{MAPE} & \textbf{MSE} & \textbf{SMAPE} & \textbf{MAPE} \\
        \hline 
        CommPre  &  $1.7676 \times 10^{-8}$ & 0.3054  &  0.2329  &  $1.6995 \times 10^{-8}$ & 0.3062 & \textbf{0.2123} \\
        \hline
        WiGesture \cite{CSI-BERT} & 3.8399 & 0.1645 & 0.1450   & 5.5027  & 0.2075 & 0.1678 \\
        \hline
        WiFall \cite{KNN-MMD} & 5.4054 & 0.1573 & 0.1434     & 5.2610 & 0.1515 & 0.1382 \\
        \hline
        WiCount  &  5.9274 & 0.1843 & \textbf{0.1550} & 5.7969 & 0.1829 & 0.1558 \\
        \hline
        \end{tabular}
\label{tab:pretrain}
\end{table*}

\subsubsection{Effect of Changes on BERT and CSI-BERT}
\begin{figure}[htbp]
\centering 
\includegraphics[width=0.49\textwidth]{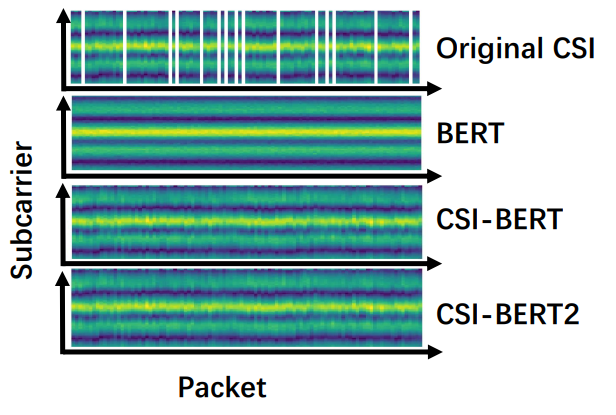}
\caption{Amplitude spectrum of original CSI and output of BERT and CSI-BERT: The blank in the original CSI represents the lost CSI.}
\label{ablation fig}
\end{figure}

To demonstrate the effectiveness of the modifications we made on BERT \cite{BERT}, we first compare the recovery performance of the original BERT (where we simply use a linear layer as the embedding and output layer to support the CSI format) with our modified version, as shown in Fig. \ref{ablation fig}. We observe that the amplitude spectrum of the original BERT appears smooth across all subcarriers, indicating that it tends to map all tokens within each subcarrier to similar value. Although BERT can also achieve a relatively low MSE of 5.26 in WiGesture dataset, it fails to capture any valuable information.

\color{black}
We then evaluate the efficacy of the introduced time embedding layer and the ARL for capturing temporal and spatial information, respectively, in comparison to CSI-BERT. The results of the ablation study for CSI classification and CSI prediction are shown in Fig. \ref{fig:pretrain} and Table \ref{tab:pretrain}, respectively.
For the CSI classification task, we observed that the time embedding layer and ARL led to significant improvements in the WiFall and WiCount datasets, but not in the WiGesture dataset. This discrepancy may be related to the difficulty of the tasks: in the WiGesture dataset, the proposed method achieved nearly 100\% accuracy even without pre-training, ARL, or the time embedding layer. In contrast, the proposed modules demonstrate greater benefits for the more complex tasks in the WiFall and WiCount datasets, , where the ARL-based spatial embedding and MLP-based time embedding resulted in improvements of approximately 13.90\% and 12.79\%, respectively.
For the CSI prediction task, both the ARL and the time embedding layer significantly reduced prediction errors across all datasets, resulting in approximately 15.47\% and 10.95\% reduction in MSE, respectively, among the four datasets. These findings illustrate that the two modules effectively help the model capture the temporal and spatial relationships within the data.

\begin{figure*}[htbp]
\centering 
\subfloat[BERT\cite{BERT} (sample set 1)]{\includegraphics[width=0.33\textwidth]{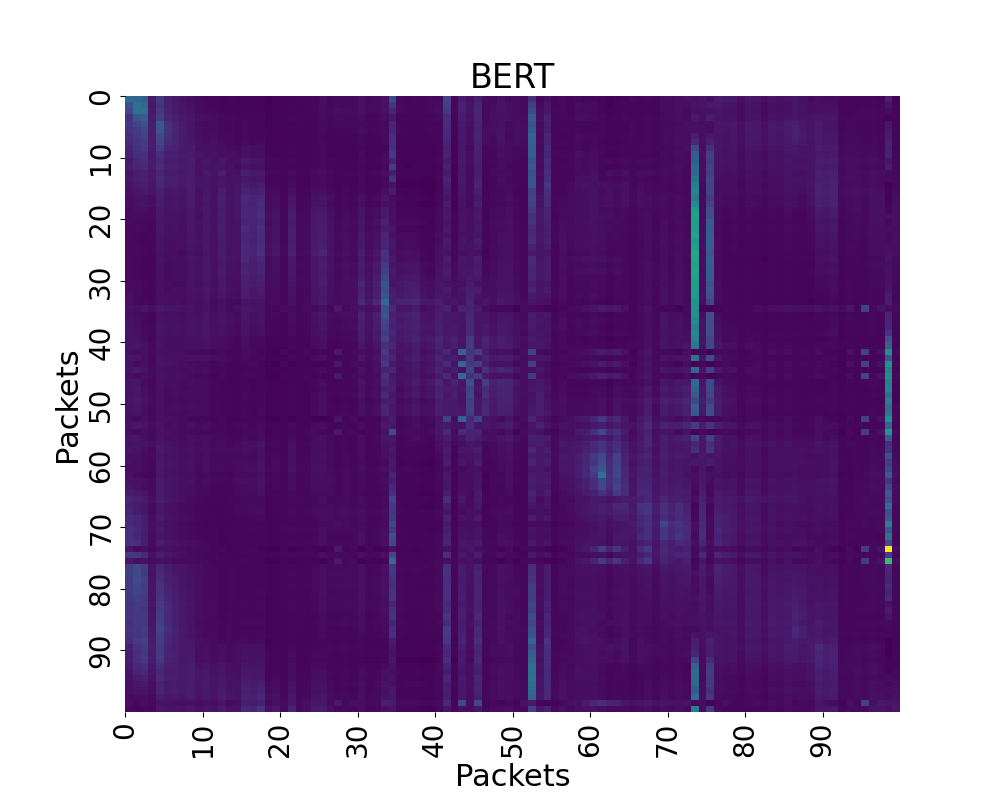}}
\subfloat[CSI-BERT\cite{CSI-BERT} (sample set 1)]{\includegraphics[width=0.33\textwidth]{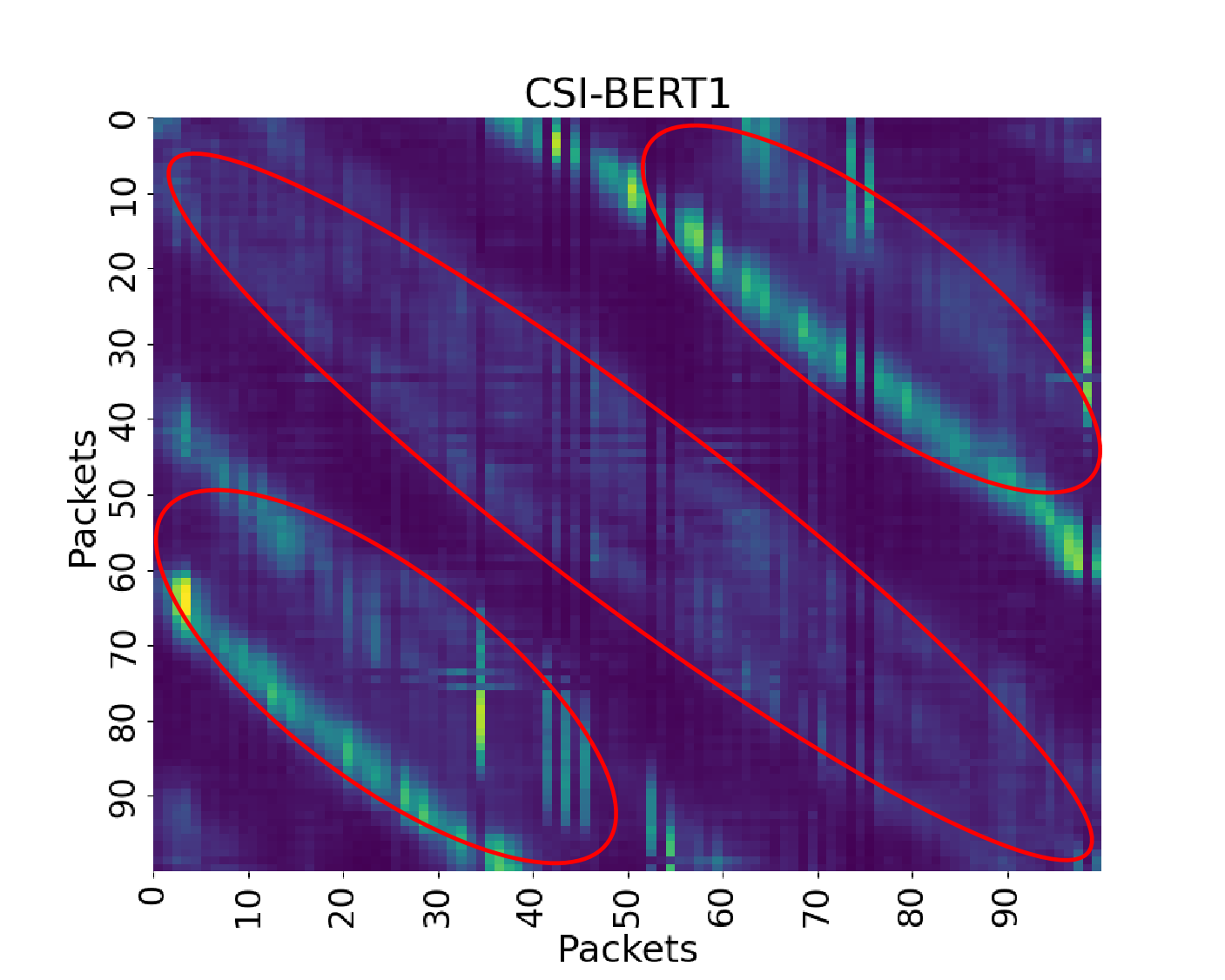}} 
\subfloat[CSI-BERT2 (sample set 1)]{\includegraphics[width=0.33\textwidth]{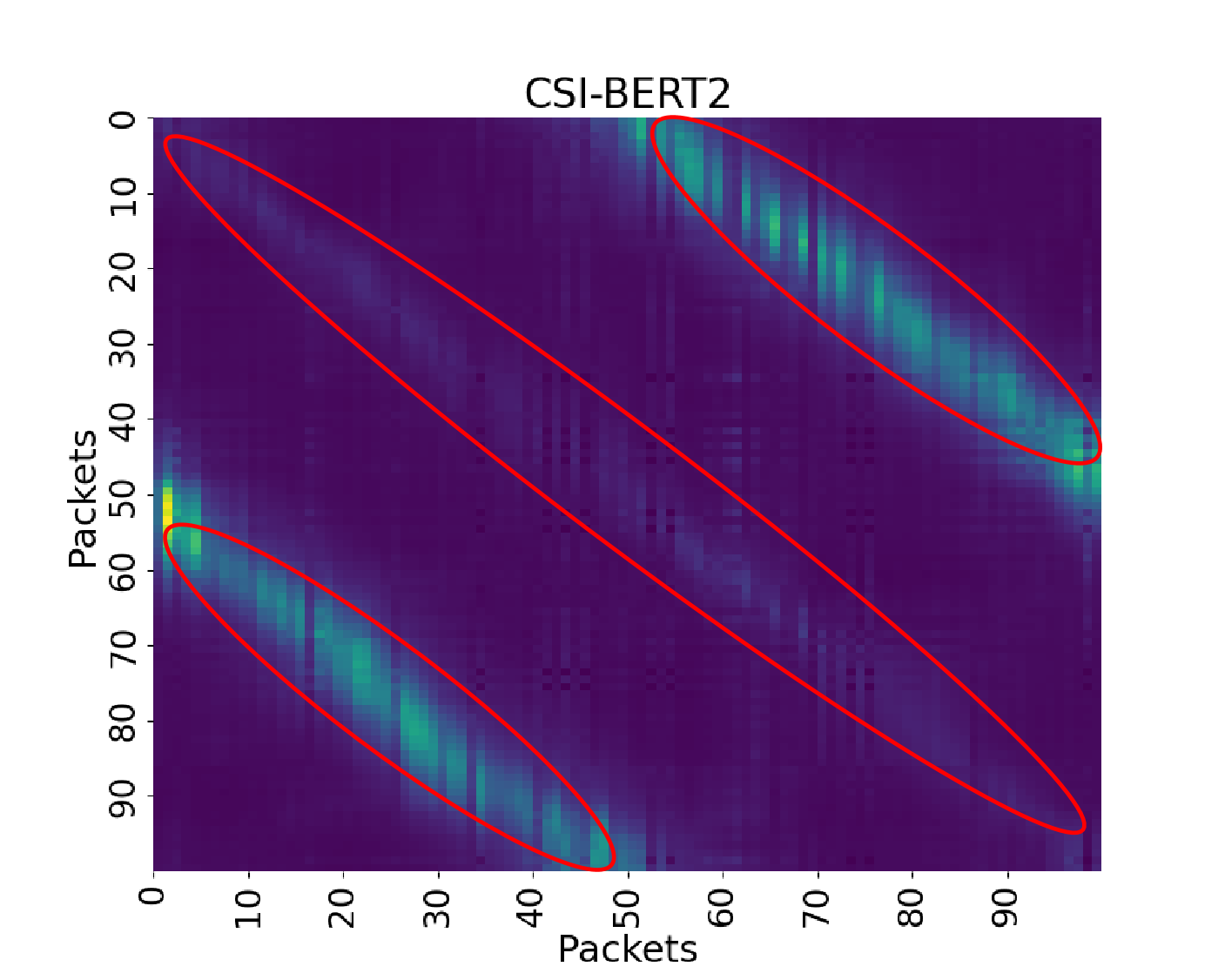}} \\
\subfloat[BERT \cite{BERT} (sample set 2)]{\includegraphics[width=0.33\textwidth]{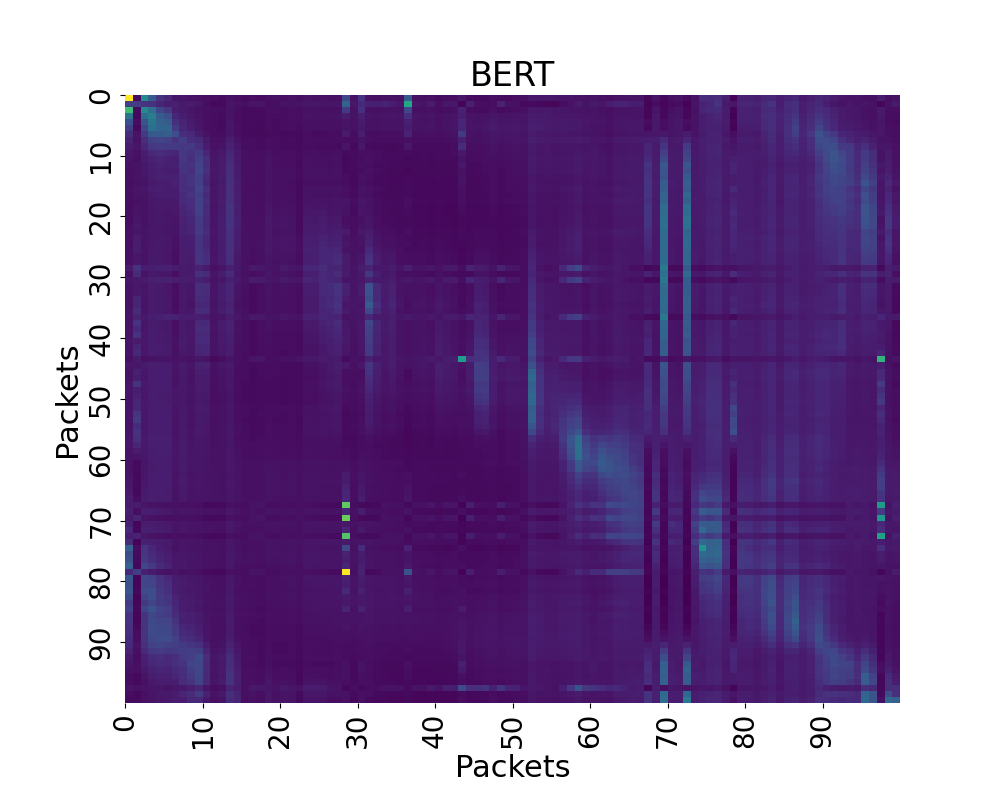}}
\subfloat[CSI-BERT\cite{CSI-BERT} (sample set 2)]{\includegraphics[width=0.33\textwidth]{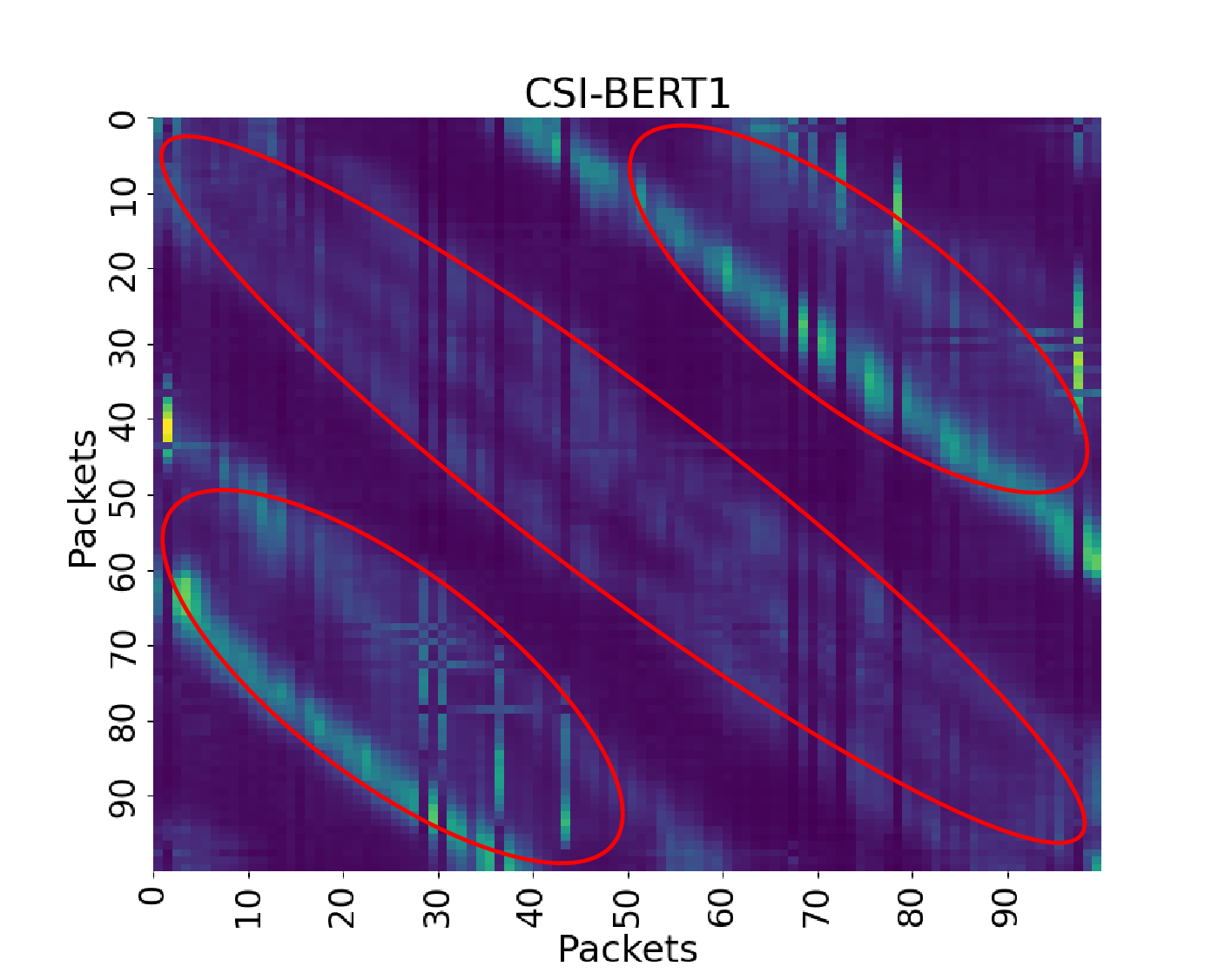}}
\subfloat[CSI-BERT2 (sample set 2)]{\includegraphics[width=0.33\textwidth]{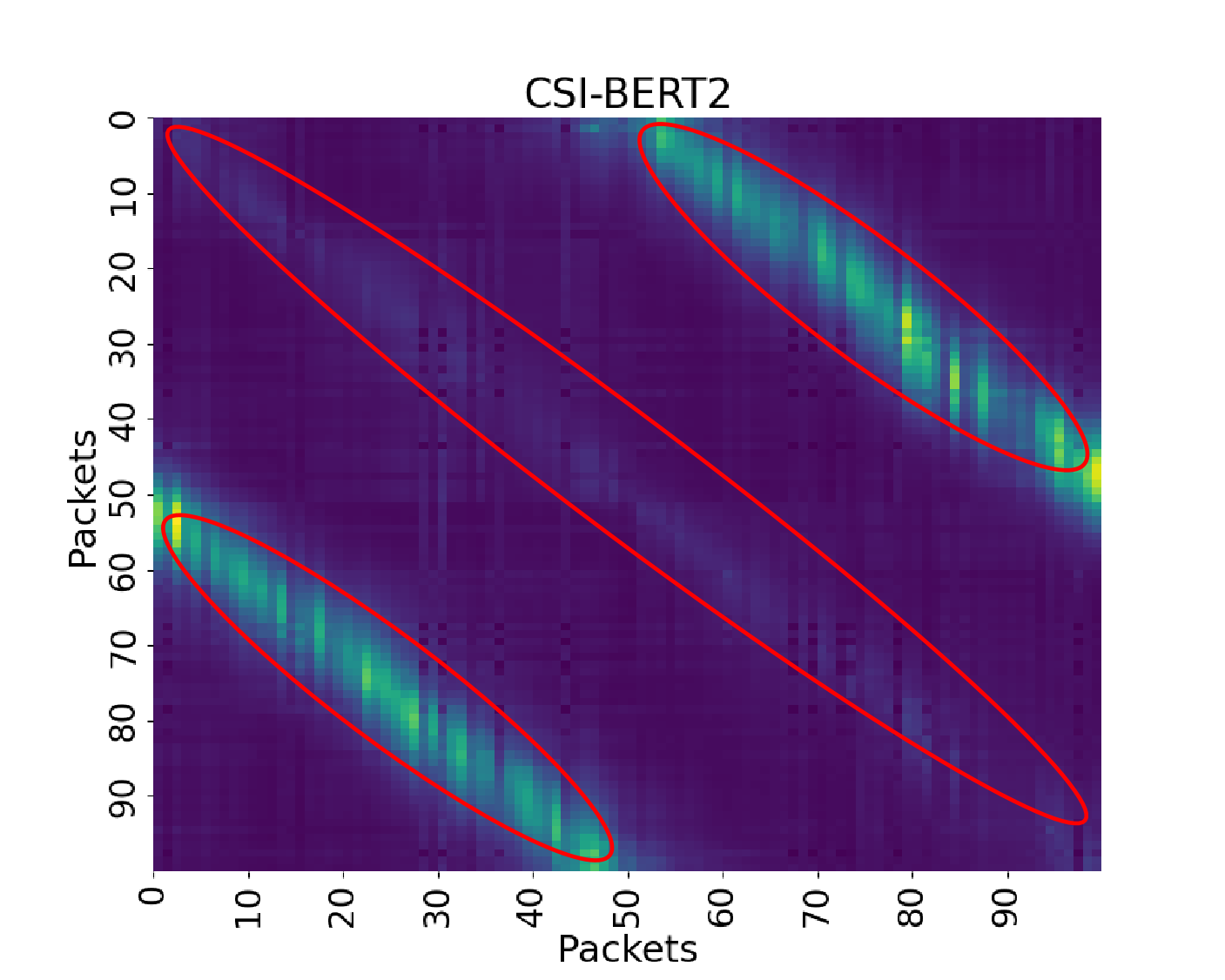}}
\caption{\textcolor{black}{Attention patterns of different models: The heatmaps illustrate the mean attention patterns of three models. The two rows show the calculation results for two different sets of 100 CSI samples respectively. Here, lighter colors represent higher values, which remains consistent across subsequent figures.}}
\label{attention pattern}
\end{figure*}

Next, we compare the attention patterns of the CSI-BERT models with those of the original BERT. Specifically, we calculate the mean value of the attention matrix from each layer:
\begin{equation}
\overline{A} = \frac{1}{LH}\sum_{l=1}^L\sum_{h=1}^H \text{Softmax}\left(\frac{{Q_{l,h}}{K_{l,h}}^T}{\sqrt{d_k}}\right) \ ,
\label{eq:attn}
\end{equation}
where ${Q_{l,h}}$ and ${K_{l,h}}$ represent the query and key of the $h$-th head in the $l$-th layer, $L$ and $H$ denote the number of layers and heads, respectively, and $d_k$ indicates their feature dimension.
\color{black}
As shown in Fig. \ref{attention pattern}, we illustrate the mean of attention patterns across all attention heads. The attention patterns of CSI-BERT and CSI-BERT2 consist of multiple diagonals from left to right, which is interpretable. This is because in a short time interval, CSI always changes in a particular regular pattern. Additionally, it is reasonable that their attention patterns are approximately symmetric, which corresponds to the mechanism of bi-directional attention in BERT. However, we cannot find any regular patterns in the attention patterns of the original BERT.

\color{black}
Furthermore, we utilize a Saliency Map to provide explanations for the CSI recovery task of CSI-BERT and BERT. We manually delete a given packet and use the models to recover it. Subsequently, we calculate the Saliency Map based on the gradient of the original input, as given by $|\frac{\partial (y-\hat{y})^2}{\partial x}|$, where $x$, $y$, and $\hat{y}$ represent the input, ground truth, and output, respectively.
\color{black}

\begin{figure*}[htbp]
\centering 
\subfloat[BERT\cite{BERT} ($20$-{th} packet)]{\includegraphics[width=0.33\textwidth]{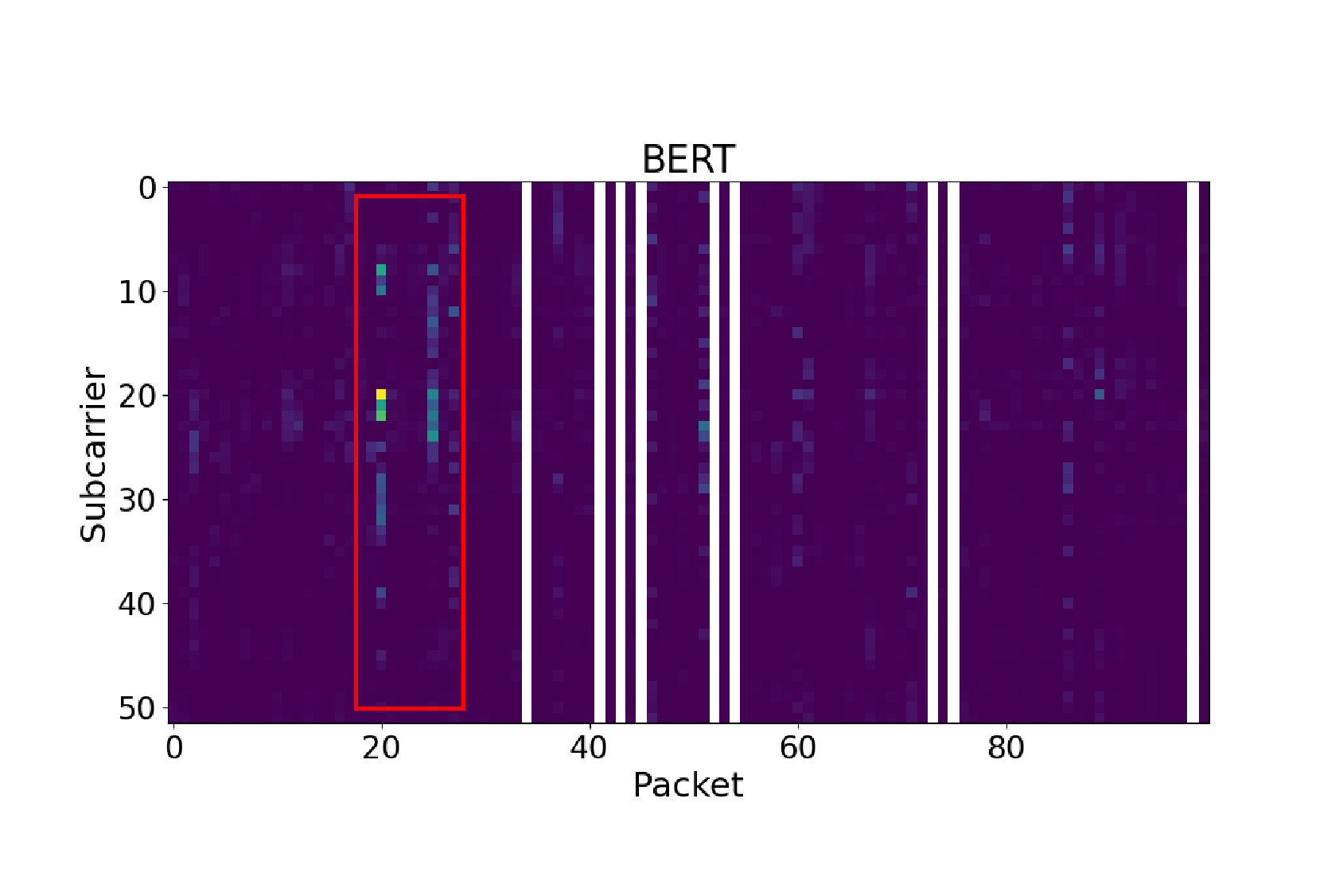}}
\subfloat[CSI-BERT\cite{CSI-BERT} ($20$-{th} packet)]{\includegraphics[width=0.33\textwidth]{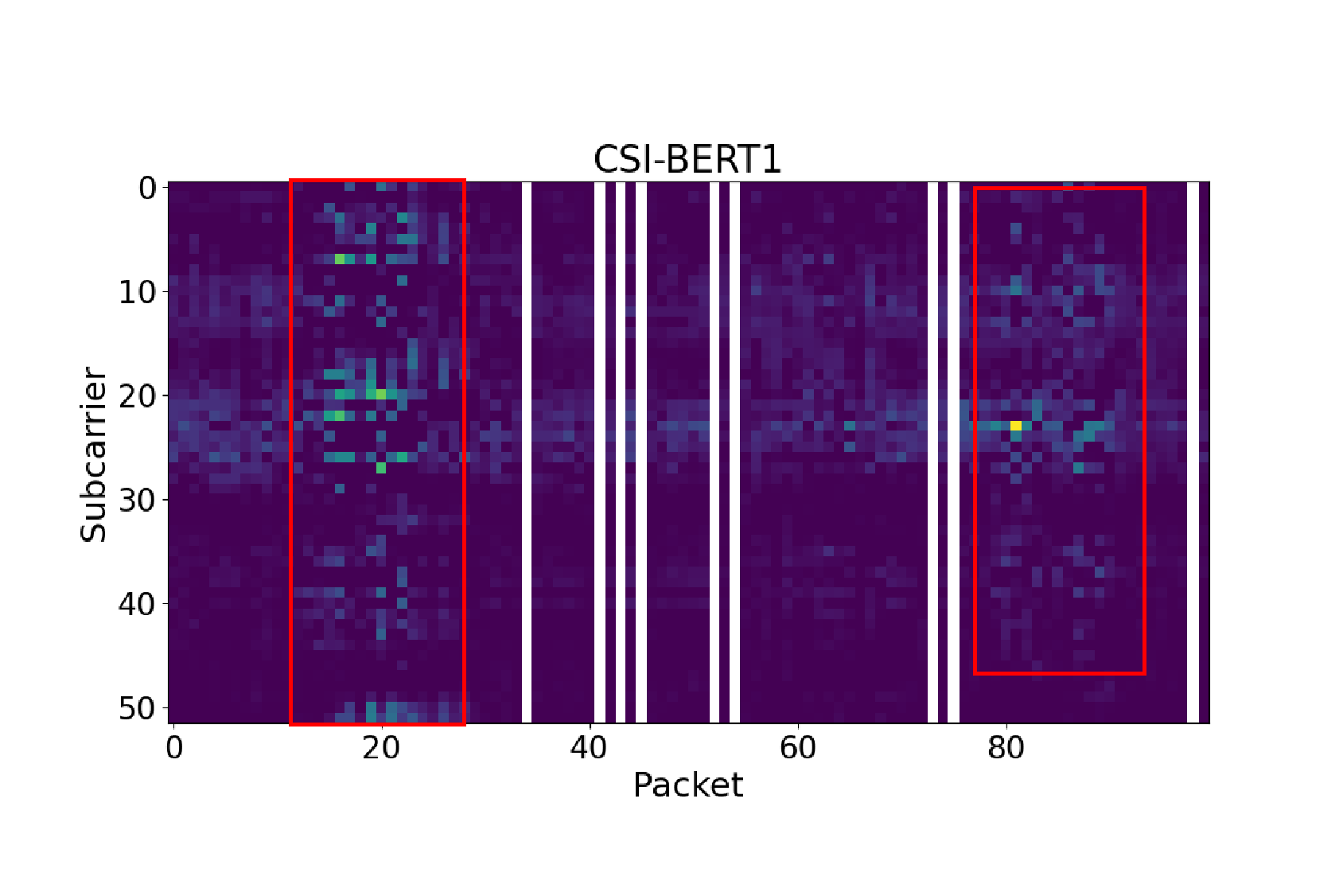}} 
\subfloat[CSI-BERT2 ($20$-{th} packet)]{\includegraphics[width=0.33\textwidth]{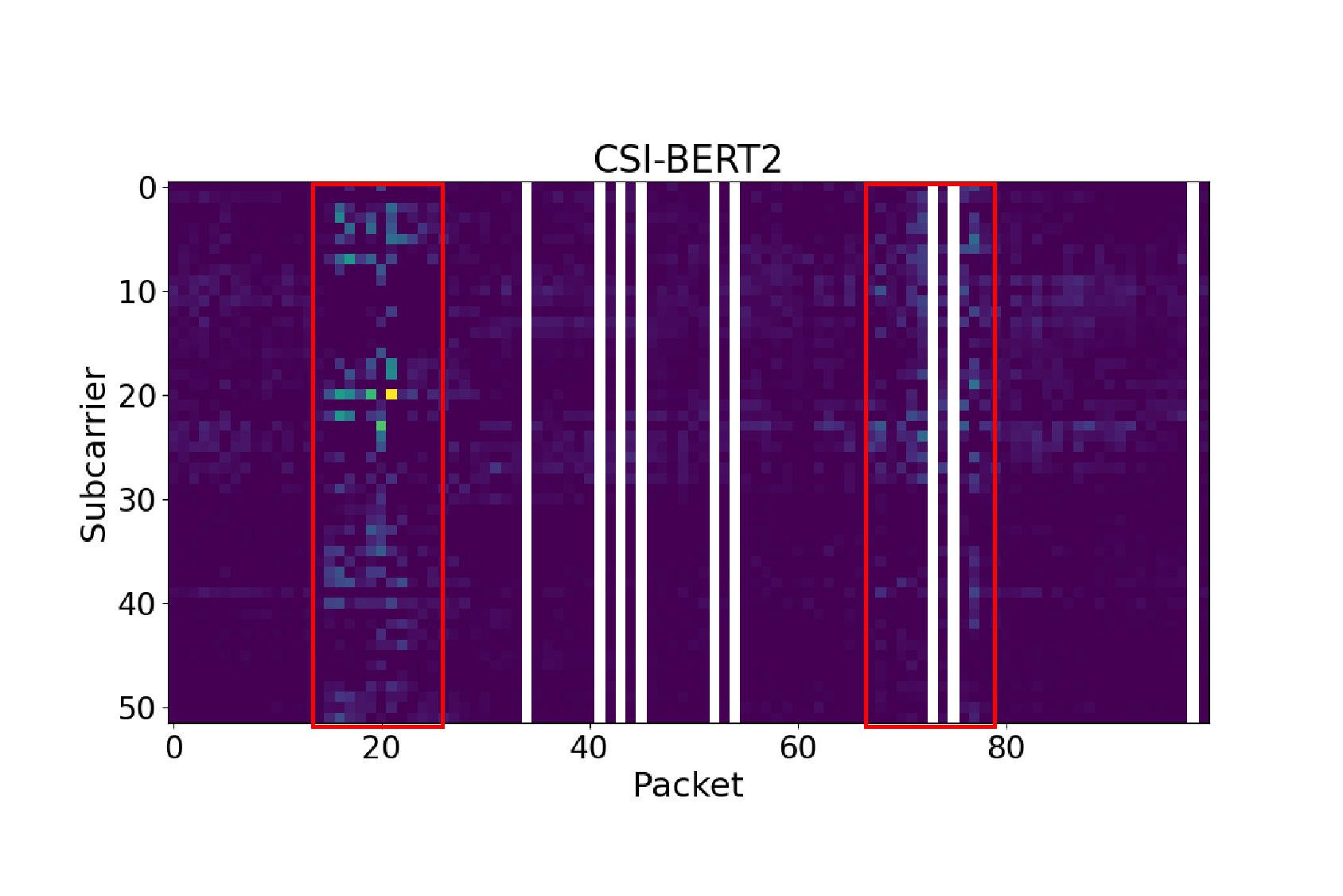}} \\
\subfloat[BERT \cite{BERT} ($80$-{th} packet)]{\includegraphics[width=0.33\textwidth]{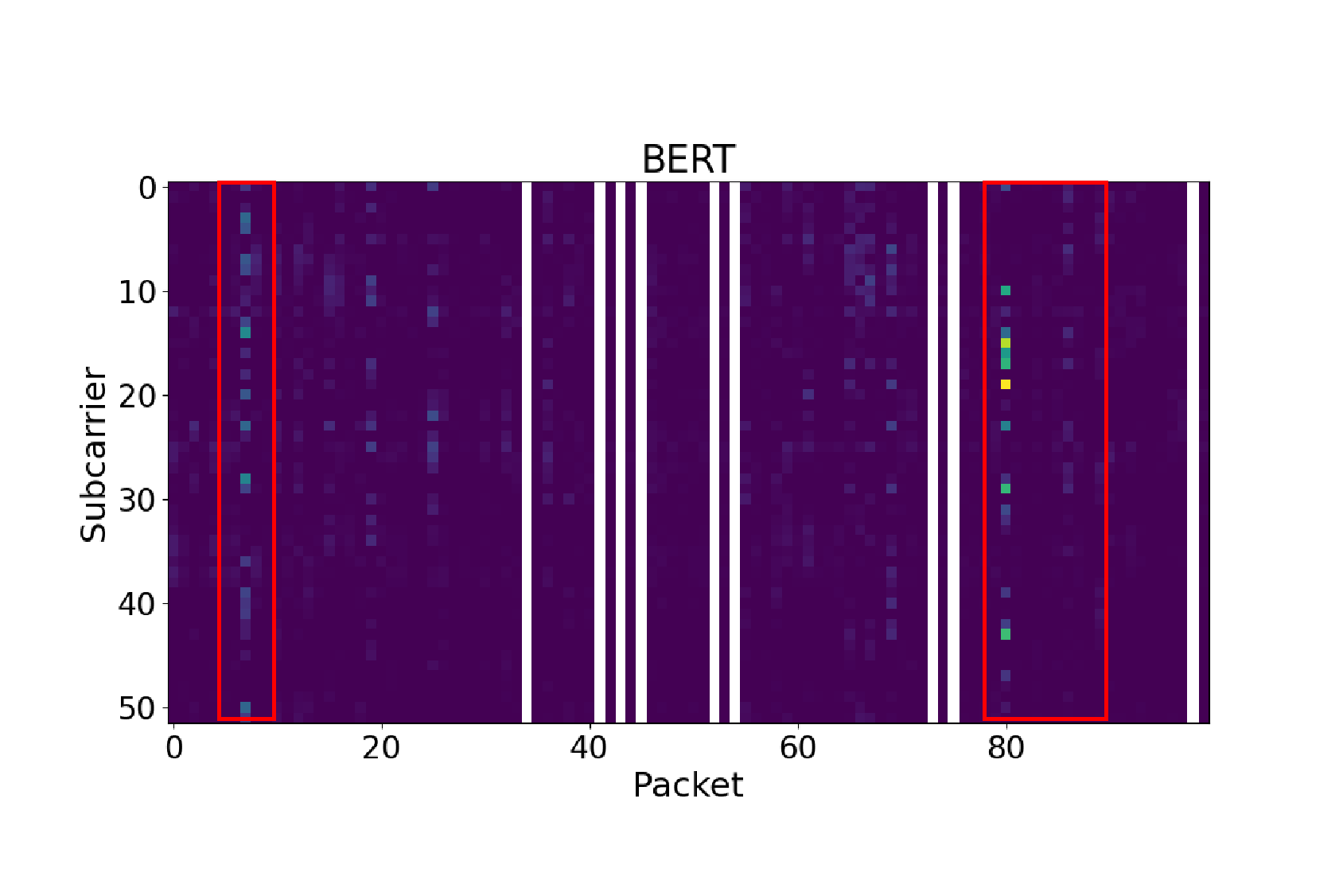}}
\subfloat[CSI-BERT \cite{CSI-BERT} ($80$-{th} packet)]{\includegraphics[width=0.33\textwidth]{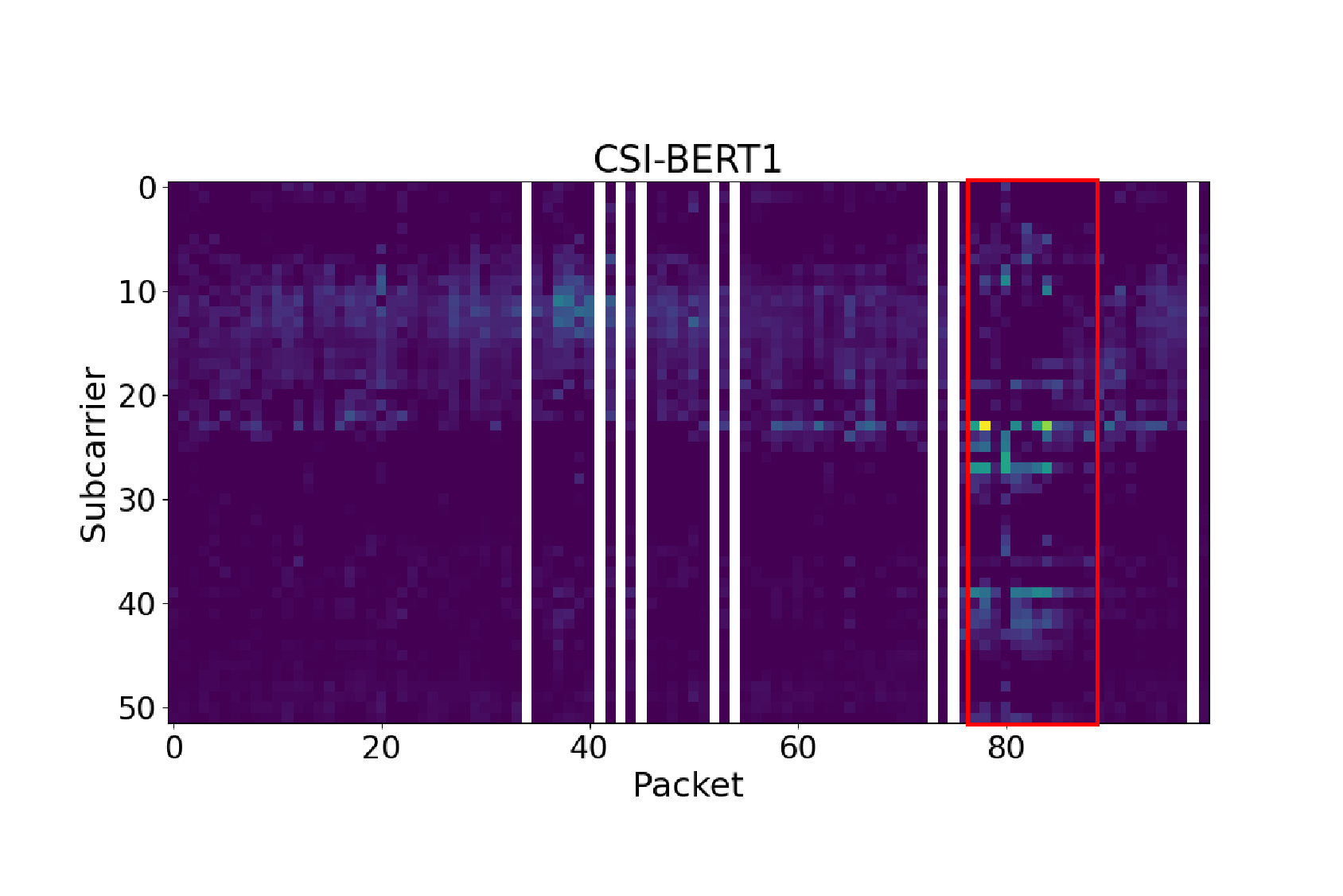}}
\subfloat[CSI-BERT2 ($80$-{th} packet)]{\includegraphics[width=0.33\textwidth]{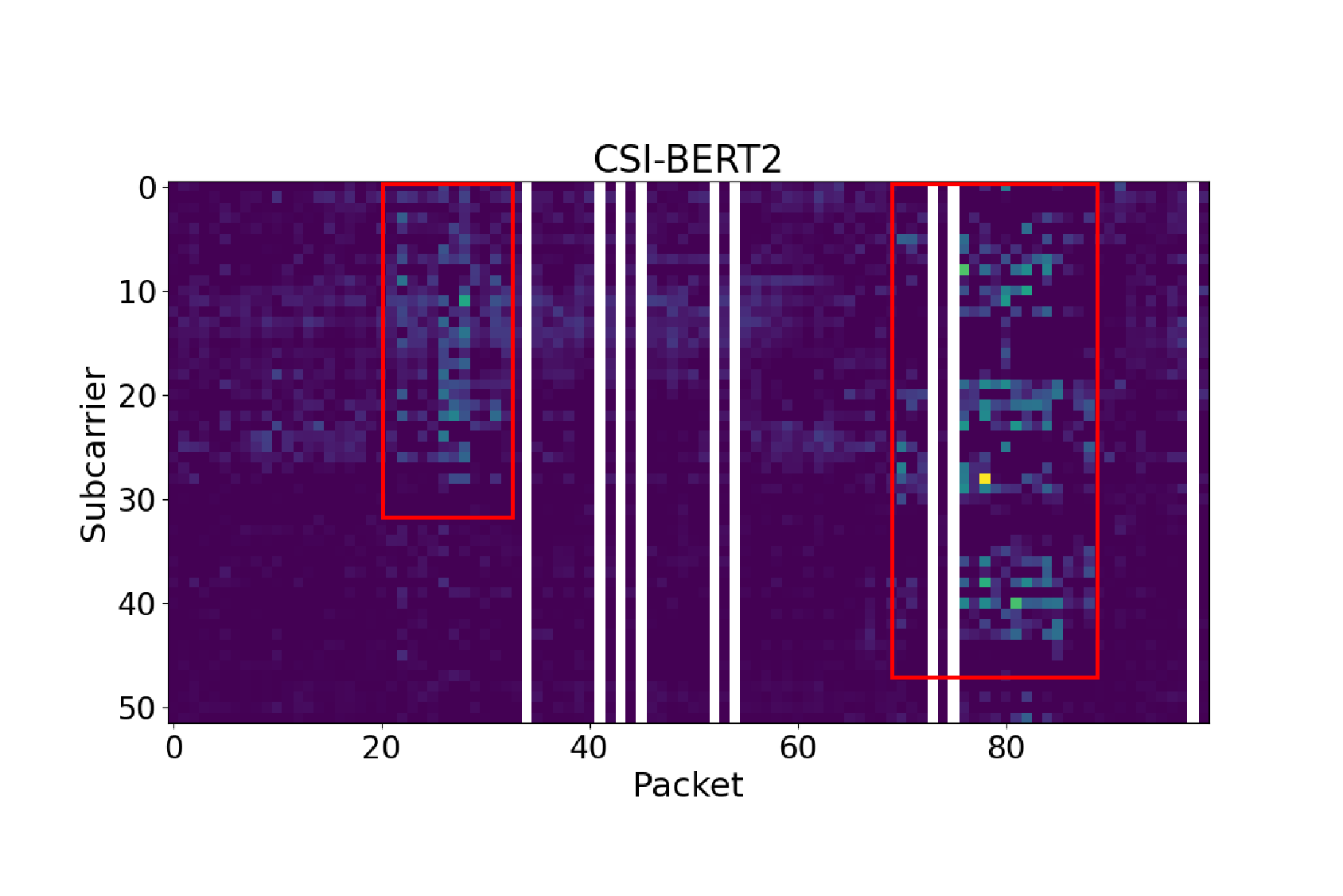}}
\caption{\textcolor{black}{Saliency Map: The two rows of heatmaps illustrate the Saliency Map when we manually delete a packet at position 20 and position 80. The blank positions represent the actual lost CSI during transmission. We calculate Saliency Map solely based on the loss function at the positions where we manually delete the packets.}}
\label{grad-cam}
\end{figure*}

We then combine it with the original input, as illustrated in Fig. \ref{grad-cam}, to visualize the regions the model primarily attends to during the recovery of the missing packet.
It can be seen that our CSI-BERT and CSI-BERT2 models mainly focus on packets nearby the lost packet, which is consistent with our intuition. However, the original BERT only focuses on the lost packet, which is actually meaningless because the lost packet is replaced by [MASK] and can hardly provide useful information, except that the packet at this position is lost.


Finally, by comparing the attention maps and Saliency Map between CSI-BERT and CSI-BERT2, we can explain why CSI-BERT2 exhibits higher generalization. In Fig. \ref{grad-cam}, we observe that CSI-BERT2 has a more centralized activation map, suggesting it utilizes less information for training while achieving better performance than CSI-BERT. This allows CSI-BERT2 to learn a simpler attention map, as shown in Fig. \ref{attention pattern}, enabling it to focus on more robust features rather than overfitting to specific relationships in the training set. This improved generalization may be attributed to the ARL and time embedding layers, which help capture more information in the dimensions of subcarrier and time (packet).


\section{Conclusion}  \label{Conclusion}
In this paper, we have proposed CSI-BERT2, a multifunctional model designed for CSI time series tasks, including both CSI prediction, and classification. Leveraging the proposed two-stage training framework, CSI-BERT2 is capable of extracting meaningful representations from limited CSI data. We have developed a novel time embedding mechanism to enhance the suitability of Transformer-based models for time series analysis. Additionally, we have employed an ARL model to improve the model's ability to capture relationships among subcarriers. Moreover, we have proposed an MPM that fine-tunes the BERT architecture specifically for CSI prediction. Furthermore, we have optimized the original training process of CSI-BERT, leading to a significant enhancement in the model's performance. These novel designs successfully address the low performance of CSI-BERT in CSI classification tasks and expand its application scenario to include CSI prediction, all within a unified framework.
Extensive experiments have validated the effectiveness of CSI-BERT2. In the CSI prediction task, it significantly outperforms traditional models while maintaining relatively fast computation speeds. In the CSI classification task, it achieves SOTA performance, even under challenging conditions where training and testing sets differ in sampling rates

\textcolor{black}{Additionally, several aspects of our method warrant further exploration in future work. First, this paper's experiments focus solely on a single-antenna and single pair TX-RX scenario. Future research could expand CSI-BERT2 to more general scenarios and explore better encoding methods to process multiple antenna inputs. Second, data heterogeneity poses a significant challenge in CSI sensing and prediction. It would be worthwhile to investigate whether CSI-BERT2 can be modified to address this challenge and serve as a foundation model, given its demonstrated potential across multiple tasks and benefits from pre-training.}



\bibliographystyle{ieeetr}
\bibliography{ref.bib}

\section*{Biography}
\begin{IEEEbiography}[{\includegraphics[width=1in,height=1.25in,clip,keepaspectratio]{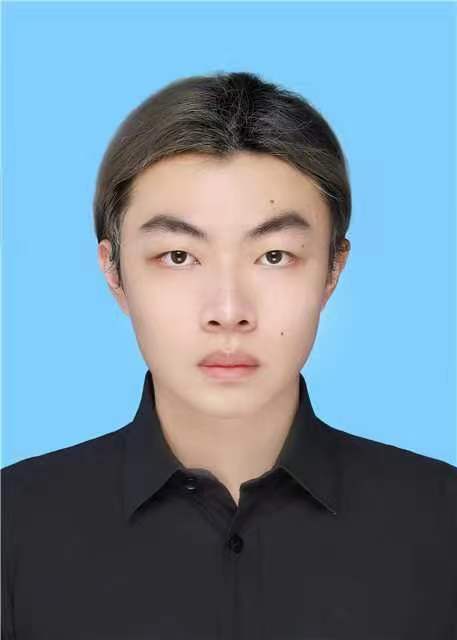}}]{Zijian Zhao}
received the B.Eng. degree in computer science and technology from School of Computer Since and Engineering, Sun Yat-sen University in 2024. He is currently pursuing the Ph.D. degree in civil engineering (scientific computation) with Department of Civil and Environmental Engineering, The Hong Kong University of Science and Technology. He was a visiting student in Shenzhen Research Institute of Big Data, The Chinese University of Hong Kong (Shenzhen) from 2023 to 2024. His current research interests include deep learning, reinforcement learning, intelligent transportation, mobile computing, and wireless sensing.
\end{IEEEbiography}

\begin{IEEEbiography}[{\includegraphics[width=1in,height=1.25in,clip,keepaspectratio]{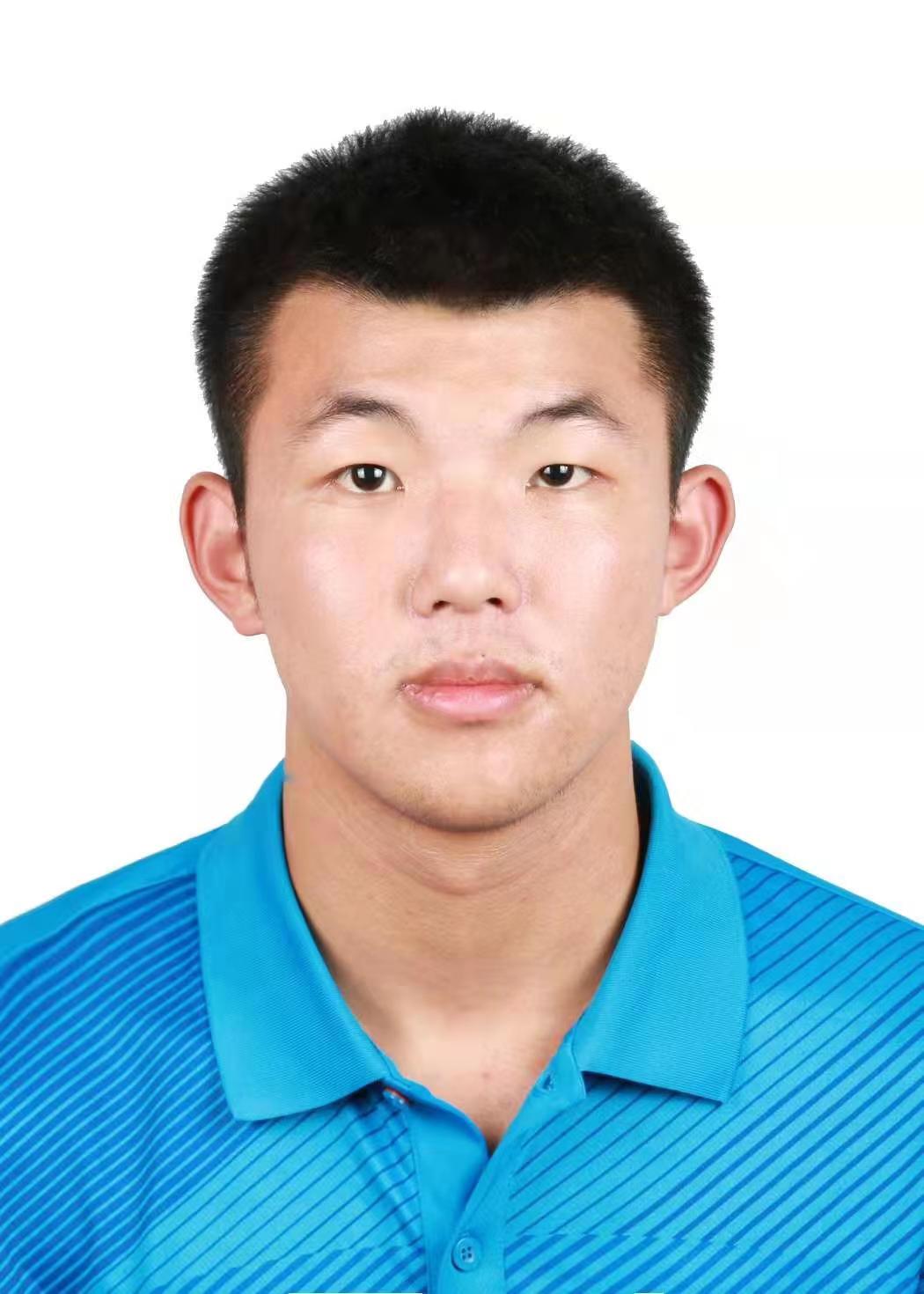}}]{Fanyi Meng}
received the B.Eng. degree in communication engineering from Southern University of Science and Technology in 2023. He is currently pursuing the M.Phil degree with Computer and Information Engineering, The Chinese University of Hong Kong (Shenzhen). His current research interests include Integrated Sensing and Communication, wireless simulation and digital twin.
\end{IEEEbiography}

\begin{IEEEbiography}[{\includegraphics[width=1in,height=1.25in,clip,keepaspectratio]{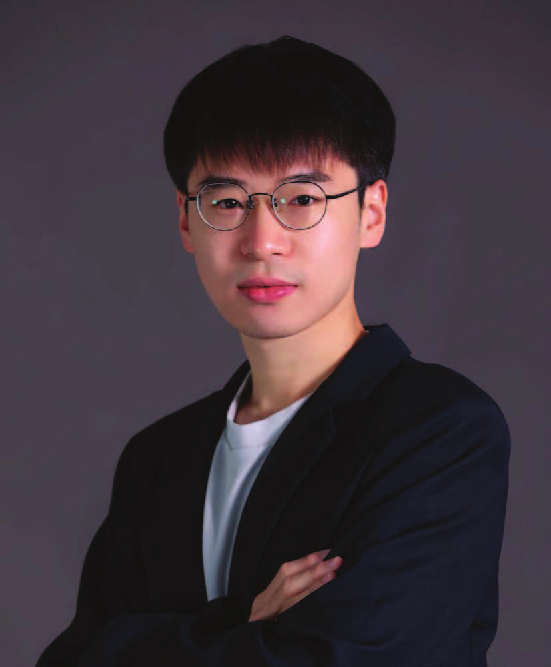}}]{Zhonghao Lyu}
received the Ph.D. degree in computer and information engineering from The Chinese University of Hong Kong, Shenzhen, China, in 2024, the M.Eng. degree in information and communication engineering from the University of Science and Technology of China, Hefei, China, in 2021, and the B.Eng. degree from Dalian University of Technology, Dalian, China, in 2018. From 2022 to 2024, he was a visiting student with the Shenzhen Research Institute of Big Data. He is currently a Post-Doctoral Research Fellow with the School of Electrical Engineering and Computer Science, KTH Royal Institute of Technology, Sweden. His research interests include large AI models, edge intelligence, and UAV communications.
\end{IEEEbiography}

\begin{IEEEbiography}[{\includegraphics[width=1in,height=1.25in,clip,keepaspectratio]{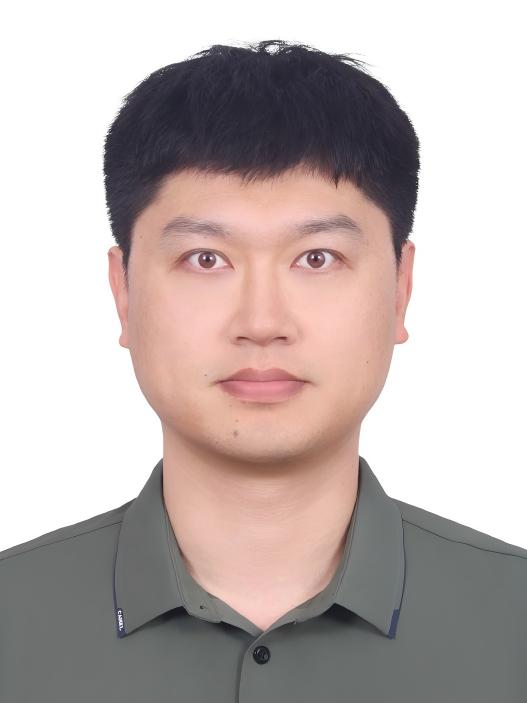}}]{Hang Li}
received the B.E. and M.S. degrees from Beihang University, Beijing, China, in 2008 and 2011, respectively, and the Ph.D. degree from Texas A\&M University, College Station, TX, USA, in 2016. He was a postdoctoral research associate with both Texas A\&M University and University of California-Davis (Sept. 2016-Mar. 2018). He was a visiting research scholar (Apr. 2018 – June 2019) and a research scientist (June 2019 – May 2025) at Shenzhen Research Institute of Big Data, Shenzhen, China. His current research interests include wireless networks, Internet of things, stochastic optimization, and applications of machine learning. He is recognized as Overseas High-Caliber Personnel (Level C) at Shenzhen in 2020.
\end{IEEEbiography}

\begin{IEEEbiography}[{\includegraphics[width=1in,height=1.25in,clip,keepaspectratio]{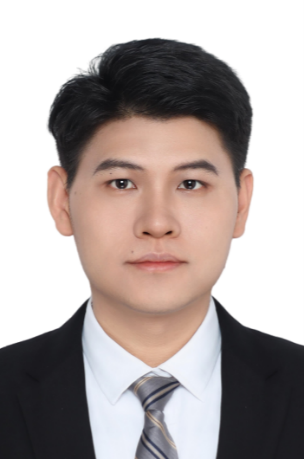}}]{Xiaoyang Li} (Member, IEEE) is currently an Assistant Professor with Southern University of Science and Technology. He received the Ph.D. degree from The University of Hong Kong. His research interests include integrated sensing-communication-computation and edge learning. He is a recipient of Young Elite Scientists Sponsorship Program by CAST, Forbes China 30 under 30, Young Elite of G20, Overseas Youth Talent in Guangdong, Overseas High-caliber Personnel in Shenzhen, Outstanding Research Fellow in Shenzhen, the Best Paper Award of IEEE 4th International Symposium on Joint Communications and Sensing, the Exemplary Reviewers of IEEE Wireless Communications Letters and Journal of Information and Intelligence (JII). He has served as the Editor of JII, and the Workshop Chairs of IEEE ICASSP/WCNC/PIMRC/MIIS/MediCom.
\end{IEEEbiography}

\begin{IEEEbiography}[{\includegraphics[width=1in,height=1.25in,clip,keepaspectratio]{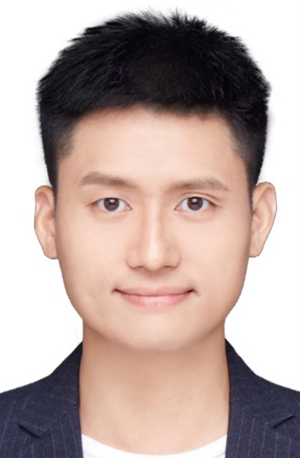}}]{Guangxu Zhu}
(Member, IEEE) received the Ph.D. degree in electrical and electronic engineering from The University of Hong Kong in 2019. Currently he is a senior research scientist and deputy director of network system optimization center at the Shenzhen research institute of big data, and an adjunct associate professor with the Chinese University of Hong Kong, Shenzhen. His recent research interests include edge intelligence, semantic communications, and integrated sensing and communication. He is a recipient of the 2023 IEEE ComSoc Asia-Pacific Best Young Researcher Award and Outstanding Paper Award, the World's Top 2\% Scientists by Stanford University, the "AI 2000 Most Influential Scholar Award Honorable Mention", the Young Scientist Award from UCOM 2023, the Best Paper Award from WCSP 2013 and IEEE  JSnC 2024. He serves as associate editors at top-tier journals in IEEE, including IEEE TMC, TWC and WCL. He is the vice co-chair of the IEEE ComSoc Asia-Pacific Board Young Professionals Committee.
\end{IEEEbiography}



 




\vfill

\end{document}